\documentclass[10pt,twocolumn,letterpaper]{article}

\usepackage[pagenumbers]{iccv} 

\usepackage{amsmath}
\usepackage{amsfonts}
\usepackage{amssymb}
\usepackage{wrapfig}
\usepackage{subcaption}
\usepackage{multirow}
 \usepackage{mathtools} 

\usepackage{verbatim}

\usepackage{anyfontsize}

\usepackage{microtype}
\usepackage{graphicx}
\usepackage{booktabs} 
\usepackage{multirow}
\usepackage{amsmath,amssymb}
\usepackage{booktabs}
\usepackage{caption,subcaption}

\usepackage{xcolor}
\definecolor{mygreen}{HTML}{3cb44b}
\definecolor{skyblue}{HTML}{beffff}
\definecolor{lightgreen}{HTML}{90ee90}

\usepackage{color, colortbl}

\definecolor{emerald}{rgb}{0.31, 0.78, 0.37}

\usepackage{tcolorbox}
\usepackage{enumitem}
\setitemize{itemsep=10pt,topsep=0pt,parsep=0pt,partopsep=0pt}
\usepackage{colortbl}

\usepackage{xcolor}
\definecolor{mygreen}{HTML}{3cb44b}
\colorlet{myyellow}{green!10!orange!90!}
\makeatletter

\usepackage{tikz}
\usetikzlibrary{arrows,shapes,snakes,automata,backgrounds,fit,petri}
\usepackage{adjustbox}

\newcommand{\RN}[1]{%
	\textup{\lowercase\expandafter{\it \romannumeral#1}}%
}
\usepackage{tabu}

\newcommand{\beq}{\vspace{0mm}\begin{equation}}
\newcommand{\eeq}{\vspace{0mm}\end{equation}}
\newcommand{\beqs}{\vspace{0mm}\begin{eqnarray}}
\newcommand{\eeqs}{\vspace{0mm}\end{eqnarray}}
\newcommand{\barr}{\begin{array}}
\newcommand{\earr}{\end{array}}

\usepackage{color, colortbl}
\definecolor{Gray}{gray}{0.93}

\usepackage{lipsum}

\usepackage{pifont}

\usepackage{makecell}

\usepackage{xcolor,amsmath}
\usepackage[linesnumbered,ruled,vlined]{algorithm2e}
\DontPrintSemicolon

\usepackage{xcolor}
\definecolor{mygreen}{HTML}{3cb44b}

\SetKwComment{Comment}{\color{green!50!black}\# }{}

\SetKwProg{Function}{def}{:}{}

\SetKwProg{For}{for}{:}{}
\SetKwProg{If}{if}{:}{}

\definecolor{iccvblue}{rgb}{0.21,0.49,0.74}
\usepackage[pagebackref,breaklinks,colorlinks,allcolors=iccvblue]{hyperref}

\usepackage{url}            
\usepackage{booktabs}       
\usepackage{amsfonts}       
\usepackage{nicefrac}       
\usepackage{microtype}      
\usepackage{xcolor}         

\usepackage{graphicx}
\usepackage{amsmath}
\usepackage{amssymb}
\usepackage{epsfig}
\usepackage{algpseudocode}
\usepackage{pifont}
\usepackage{multirow,makecell}
\usepackage{bbm}
\usepackage{mathtools}
\usepackage{diagbox}
\usepackage{enumitem}
\usepackage{soul}
\usepackage{color}
\usepackage{colortbl}
\usepackage{arydshln}
\usepackage{alltt}

\tcbuselibrary{most}
\tcbset{
  aibox/.style={
    width=\textwidth,
    top=10pt,
    colback=white,
    colframe=black,
    colbacktitle=black,
    enhanced,
    center,
    attach boxed title to top left={yshift=-0.1in,xshift=0.15in},
    boxed title style={boxrule=0pt,colframe=white,},
  }
}
\newtcolorbox{AIbox}[2][]{aibox,title=#2,#1}
\newlength\savewidth

\def\confName{ICCV}
\def\confYear{2025}

\title{EVEv2: Improved Baselines for Encoder-Free Vision-Language Models}

\newcommand*{\affaddr}[1]{#1} 
\newcommand*{\affmark}[1][*]{\textsuperscript{#1}}

\author{
Haiwen Diao\textsuperscript{1,2}\thanks{Equal contribution. $\dag$ Correspondence to \textit{wangxinlong@baai.ac.cn}.}\hspace{2.7mm}
Xiaotong Li\affmark[3,2]$^*$\hspace{1.5mm}
Yufeng Cui\affmark[2]$^*$\hspace{1.5mm}
Yueze Wang\affmark[2]$^*$\hspace{1.5mm}\\
Haoge Deng\affmark[4,2]\hspace{1.5mm}
Ting Pan\affmark[5,2]\hspace{1.5mm}
Wenxuan Wang\affmark[6,5,2]\hspace{1.5mm}
Huchuan Lu\affmark[1\dag]\hspace{1.5mm} 
Xinlong Wang\affmark[2\dag]\hspace{1.5mm}\\
\affaddr{
\affmark[1]DLUT\hspace{1.5mm}
\affmark[2]BAAI\hspace{1.5mm}
\affmark[3]PKU\hspace{1.5mm}
\affmark[4]BUPT\hspace{1.5mm}
\affmark[5]UCAS\hspace{1.5mm}
\affmark[6]CASIA}
}

\begin{document}
\maketitle
\begin{abstract}
Existing encoder-free vision-language models (VLMs) are rapidly narrowing the performance gap with their encoder-based counterparts, highlighting the promising potential for unified multimodal systems with structural simplicity and efficient deployment. 
We systematically clarify the performance gap between VLMs using pre-trained vision encoders, discrete tokenizers, and minimalist visual layers from scratch, deeply excavating the under-examined characteristics of encoder-free VLMs. We develop efficient strategies for encoder-free VLMs that rival mainstream encoder-based ones.
After an in-depth investigation, we launch \textbf{EVEv2.0}, a new and improved family of encoder-free VLMs.
We show that: 
(i) Properly decomposing and hierarchically associating vision and language within a unified model reduces interference between modalities.
(ii) A well-designed training strategy enables effective optimization for encoder-free VLMs.
Through extensive evaluation, our \textbf{EVEv2.0} represents a thorough study for developing a decoder-only architecture across modalities, demonstrating superior data efficiency and strong vision-reasoning capability.
Code is publicly available at: \href{https://github.com/baaivision/EVE}{\textcolor{blue}{https://github.com/baaivision/EVE}}.
\end{abstract}    
\section{Introduction}
\label{sec:Introduction}

\begin{figure*}[t]
    \vspace{-2mm}
    \centering 
    \includegraphics[width=0.98\linewidth,trim= 0 0 0 0,clip]
    {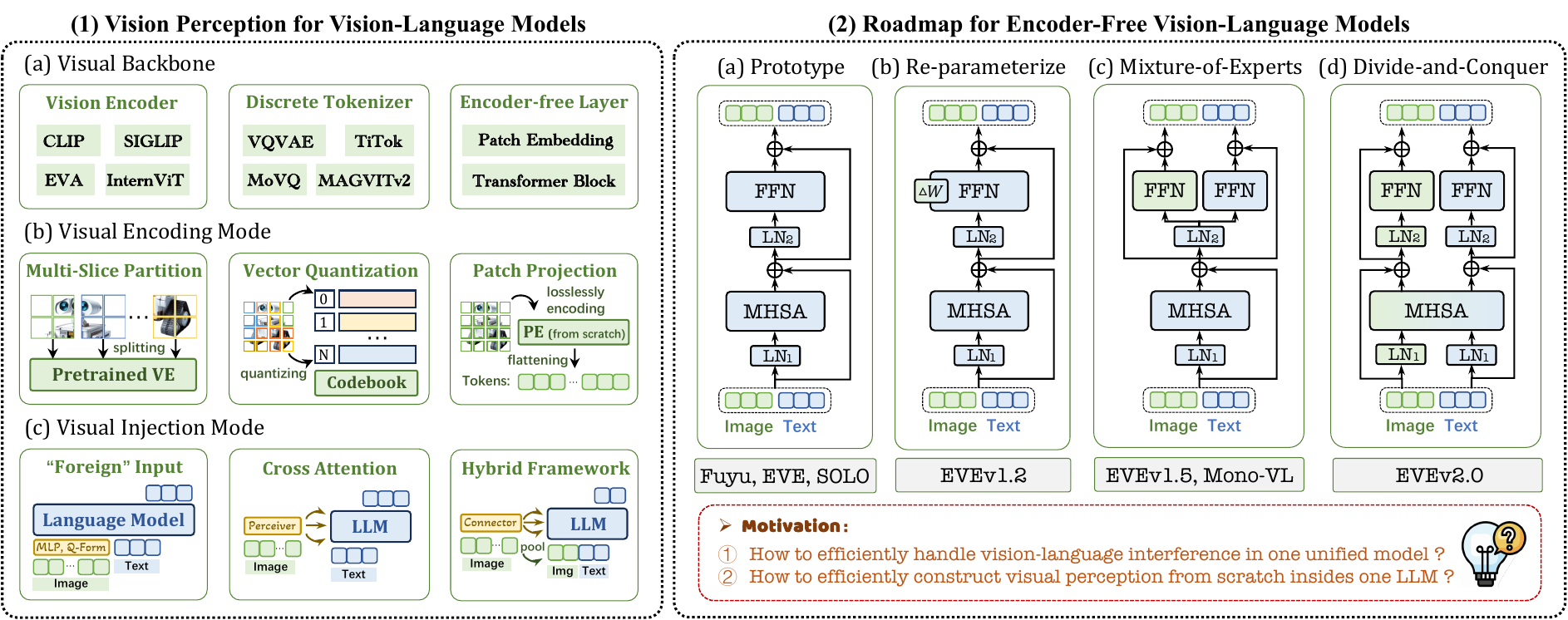} 
    \caption{Overview of (1) diverse vision construction inside existing VLMs and (2) potential architecture variants of Encoder-Free VLMs.}
    \label{fig:motivation}
\end{figure*}

With the recent rapid advancements in the large language models (LLMs)~\cite{VLM:GPT-4,TransF:LLaMA2,yang2024qwen2,TransF:InternLM2,TransF:Deepseekllm} and large vision models (LVMs)~\cite{TransF:ViT,TransF:EVA,TransF:EVA-CLIP-18B,TransF:ViT-22B,VLM:mc-beit}, vision-language models (VLMs)~\cite{VLM:qwen-vl,VLM:DeepSeek-VL,VLM:GPT-4v,VLM:Claude3,TL:UniPT,TL:SHERL} have made remarkable strides, demonstrating impressive capabilities in multi-modal understanding and reasoning applications.
As illustrated in~\Cref{fig:motivation}(1), typical practice adopts pre-trained vision encoders to extract visual semantics, which are then translated into the text embedding space as ``Foreign'' language inputs for subsequent LLMs (\eg, BLIP~\cite{VLP:BLIPv2} and LLaVA~\cite{VLM:LLaVA}). 
In contrast, another representative branch transforms visual features from vision encoders' last layer across each layer of LLMs through cross-attention modules, like Flamingo~\cite{VLP:Flamingo} and LLaMA-3.2V~\cite{VLMs:LLama3.2}.
Thanks to well-aligned representations across modalities through large-scale contrastive learning~\cite{VLP:CLIP,TransF:Siglip,TransF:EVA-CLIP}, these studies can achieve promising performance and strong multi-modality capability after joint training. 
However, the inductive biases during visual pre-training, \eg, image resolution, aspect ratio, and semantic priors, limit the flexibility and applicability of the visual learning in diverse real-world scenarios~\cite{VLM:LLaVA-UHD,tong2024eyes,tong2024cambrian}. 

Unlike them, Fuyu~\cite{VLM:Fuyu-8b} takes an early step to explore monolithic VLMs via removing pre-trained visual encoders, while EVE~\cite{VLM:EVE} first pioneers a transparent, efficient, and practical path for advancing the encoder-free VLM direction.
The subsequent PaliGemma~\cite{beyer2024paligemma} constructs an encoder-free VLM that shows strong scaling efficiency while progressively approaching its encoder-based counterpart.
With extensive training data and computing resources, Mono-InternVL~\cite{luo2024mono} narrows the gap and matches the performance of Intern-VL1.5~\cite{VLM:InternVL-1.5}, starting with the same LLM capabilities.

\begin{figure*}[t]
    \centering 
    \includegraphics[width=0.96\linewidth,trim= 0 0 0 0,clip]
    {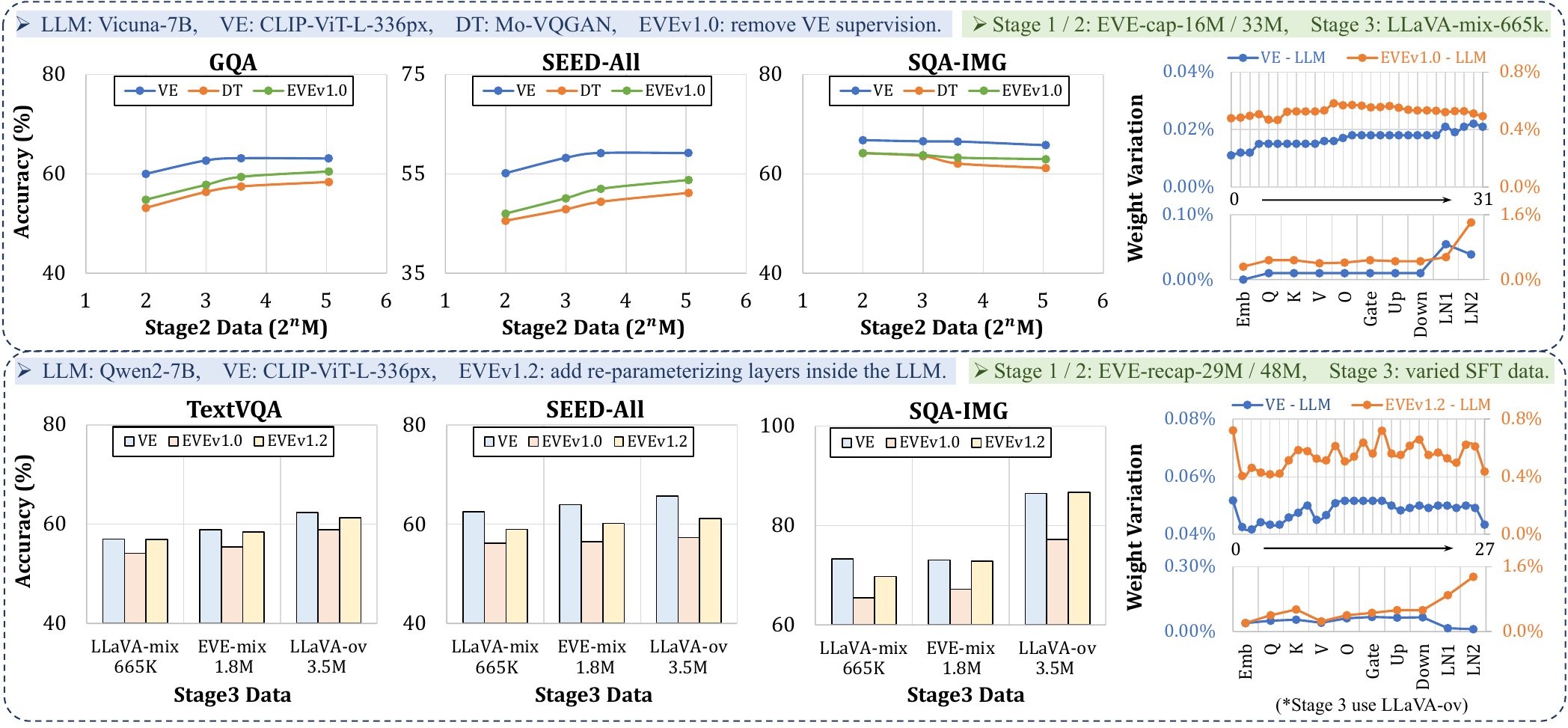} 
    \caption{Preliminary scaling efficiency analyses during pre-training or fine-tuning across various VLMs. (More details in the Appendix). 
    Notably, VE / DT / EVE apply varying image downsampling rates (14$^2$ / 8$^2$ / 32$^2$). For fairness, we choose different resolutions that yield relatively balanced token counts of 576 / 1024 / 625 tokens per image. Besides, we quantify weight changes between LLMs and VLMs by averaging absolute value variation within specific layer number or type. We report GQA~\cite{Datasets:GQA}, SEED~\cite{VLM:SEED}, and TextVQA~\cite{Datasets:TextVQA} for in-domain, open-domain, and OCR-related validation. 
    \textit{Note that SQA~\cite{Datasets:ScienceQA} involves text-related knowledge tasks susceptible to LLM's forgetting issue.}
    }
    \label{fig:preliminary}
\end{figure*}

However, constructing encoder-free VLMs remains challenging, particularly in learning vision perception from scratch and reducing vision-language interference within a unified model.
To date, three solutions have been put forward:
(i) Visual feature supervision~\cite{VLM:EVE}; 
(ii) Incremental training recipes~\cite{VLM:EVE,chen2024solo,luo2024mono}; 
(iii) Mixture-of-Expert (MoE) detachment~\cite{lin2024moma,luo2024mono} in~\Cref{fig:motivation}(2c).
Nevertheless, we empirically discover that visual supervision can be substituted by large-scale, high-quality image-text datasets annotated by a powerful captioning engine.
During training, properly merging language and multimodal data helps mitigate knowledge-forgetting issues, while pressuring the development of multimodal capabilities.
From the view of VLM structure, decoupling partial visual functions from the unified model through a MoE design~\cite{bao2022vlmo,VLP:BEiTv3,li2024aria} aids in relieving vision-language interference to some extent. 
However, we discover significant weight shifts across various network layers between the VLMs and the original LLMs, revealing the insufficiency of the current decoupling degree. Notably, we further exploit a re-parameterize architecture in~\Cref{fig:motivation}(2b) for seamless LLM adaptation. Although yielding better gains over prototype EVE in~\Cref{fig:preliminary}(2a), it does not completely resolve the representational conflicts across modalities.

From the above observations, we launch \textbf{EVEv2.0}, a new and improved baseline for encoder-free VLMs.
\textbf{EVEv2.0} completely disentangles overall components and introduces modality-wise sparsity into one unified decoder-only backbone in~\Cref{fig:motivation}(2d). Such a Divide-and-Conquer architecture maximizes scaling efficiency in building visual perception while minimizing the negative influence on the LLM itself.
Besides, using an enhanced caption engine, we construct an effective and practical route for monolithic VLM research that facilitates data-scaling efficiency and stably transfers increasingly stronger LLMs into the encoder-free VLMs.
With 100M publicly available data, \textbf{EVEv2.0} outperforms encoder-free counterparts, and continually approaches encoder-based competitors of similar capacity across diverse vision-language benchmarks. 
Our \textbf{EVEv2.0} unveils valuable insights for developing scalable, native, and next-gen VLMs, paving a transparent roadmap for future research supported by larger training data and computational resources.

\section{Related Work}
\label{sec:Related-work}

\textbf{Encoder-based VLMs.} 
Encoder-based methods have become the dominant approach in vision-language models, widely adopted in commercial products, \eg. GPT-4V~\cite{VLM:GPT-4}, Claude 3.5~\cite{VLM:Claude3}, and Gemini~\cite{VLM:Gemini}, as well as in the open-source projects like LLaVA series~\cite{VLM:LLaVA, VLM:LLaVA-1.5, VLM:LLaVA-1.6, Llava-onevision}, Qwen-VL series~\cite{VLM:qwen-vl,wang2024qwen2-vl}, InternVL series~\cite{VLM:InternVL, VLM:InternVL-1.5}, BLIP series~\cite{VLP:BLIP,VLP:BLIPv2,xue2024blip-3}, and EMU series~\cite{VLM:EMU,VLM:EMUv2}. They benefit from the pre-trained knowledge from visual encoders~\cite{VLP:CLIP,TransF:Siglip,TransF:EVA-CLIP,CL:DINO} and LLMs~\cite{TransF:LLaMA,TransF:LLaMA2,TransF:Vicuna,TransF:InternLM,TransF:InternLM2,TransF:Qwen,yang2024qwen2,qwen2.5}, successfully building modular VLMs for various real-world applications. Among them, most studies~\cite{VLM:mPLUG-Owl,VLM:mPLUG-Owl2,VLM:Monkey,VLM:LLaVA-UHD,VLM:Sharegpt4v,VLM:Densefusion} directly translate vision representations into the input space of LLMs.
In contrast, another type of research~\cite{VLM:Llama-Adapter,VLM:Llama-Adapterv2,VLM:CogAgent,VLP:Flamingo,VLMs:LLama3.2} introduces the cross-attention module for integrating visual and language information layer-by-layer.
Despite achieving strong performance gains, it may be insufficient to simply map visual information into the input space of LLMs~\cite{VLM:LLaVA-1.5,VLM:mPLUG-Owl2} or connect the same visual features across different representational levels of the LLM~\cite{VLM:CogAgent,VLM:IDEFICS}, given the heterogeneous characteristics between vision and language. 
Besides, these modular VLMs face challenges in further development due to the strong inductive biases in pre-training visual encoding patterns, complex infrastructure requirements, and scaling laws necessary to balance various separate components.

\noindent\textbf{Encoder-free VLMs.} 
Another visual processing strategy is discrete visual tokenization~\cite{van2017VQVAE,VQGAN,LLamagen}, which translates the image into a sequence of discrete tokens and is widely used in various multi-modal understanding and generation approaches~\cite{team2024chameleon,VLM:AnyGPT,wang2024mio,VLM:EMU3}.
However, the discretization inevitably results in lossy visual information and weakens in extracting semantic contents, which in turn hinders fine-grained visual understanding and reasoning~\cite{xie2024show-o,wu2024janus}.
Therefore, recent studies~\cite{wu2024vila-u,li2024imagefolder,qu2024tokenflow,xie2024muse} introduce semantic constraints in the visual tokenizer for both high-level semantic and fine-grained visual representations.
Compared to their highly-compressed features in the discrete-value space, encoder-free VLMs~\cite{VLM:Fuyu-8b,VLM:EVE,chen2024solo} have emerged as a promising architecture for lossless visual encoding, efficient data scaling, and end-to-end multimodal processing. 
Specially, EVE~\cite{VLM:EVE} pioneers an efficient and transparent path for monolithic VLMs, with its data-scaling efficiency preliminarily validated by PaliGemma~\cite{beyer2024paligemma}. Impressively, Mono-InternVL~\cite{luo2024mono} bridges the performance gap with its modular counterpart~\cite{VLM:InternVL-1.5} of the same LLM capacity, using adequate data. We emphasize that limited by current training data and device resources, EVEv2.0 does not aim for state-of-the-art performance but instead focuses on revealing the most efficient route for encoder-free VLMs from scratch.

\section{Methodology}
\label{sec:Methodology}

\subsection{Preliminary}
\label{subsec:preliminary}

\textbf{Settings.} Building on~\cite{VLM:LLaVA-1.5,VLM:EMU3,VLM:EVE}, we conduct two pilot experiments in~\Cref{fig:preliminary}. 
\textbf{Exp.(i):} we combine Vicuna-7B~\cite{TransF:Vicuna} with vision encoder (\textit{VE}: CLIP-ViT-L-336px~\cite{VLP:CLIP}), discrete tokenizer (\textit{DT}: Mo-VQGAN~\cite{VLM:movq}), or one single transformer block~\cite{TransF:Transformer} (EVEv1.0: w/o VE supervision). 
We first train the projector, vision vocabulary, or vision block using EVE-cap-16M~\cite{VLM:EVE} caption data, followed by EVE-cap-33M~\cite{VLM:EVE} to update only vision encoder for \textit{VE} (work best), or extra LLM weights for \textit{DT} and EVEv1.0. 
Finally, we update all layers except for discrete tokenizers during fine-tuning. 
\textbf{Exp.(ii):} we use stronger Qwen2-7B~\cite{TransF:Qwen} as the LLM for \textit{VE}, EVEv1.0, and EVEv1.2.
Using 29M image-text pairs, we initially train the projector for \textit{VE} or all visual-related layers for EVE, with further updates to EVE’s LLM layers via an extra 48M data. 
We then import varying-scale instruction data~\cite{VLM:LLaVA-1.5,VLM:EVE,Llava-onevision} to train the entire network.

\textbf{Finding (1):}
\textit{Performance gap between various vision encoding modes.} Exp.(i) shows that initially, \textit{VE} performs better than \textit{DT} and EVEv1.0 in visual understanding due to much larger image-text pretraining datasets (400M) and already alignment space between vision-language embeddings.
Despite building visual recognition from scratch, EVEv1.0 demonstrates strong scaling properties,
progressively closing the performance gap with \textit{VE} as the data scale increases. Subsequent studies~\cite{beyer2024paligemma,luo2024mono} have proved that, with sufficient data, they can achieve comparable performance.
Notably, \textit{DT} maps visual information into a discrete space through quantization, hampering effective visual perception and weakening vision-language association via an image reconstruction objective. This ultimately results in subpar performance, even at larger data scales, leaving \textit{DT} less competitive overall.

\begin{figure*}[t]
    \centering 
    \includegraphics[width=0.98\linewidth,trim= 0 0 0 0,clip]
    {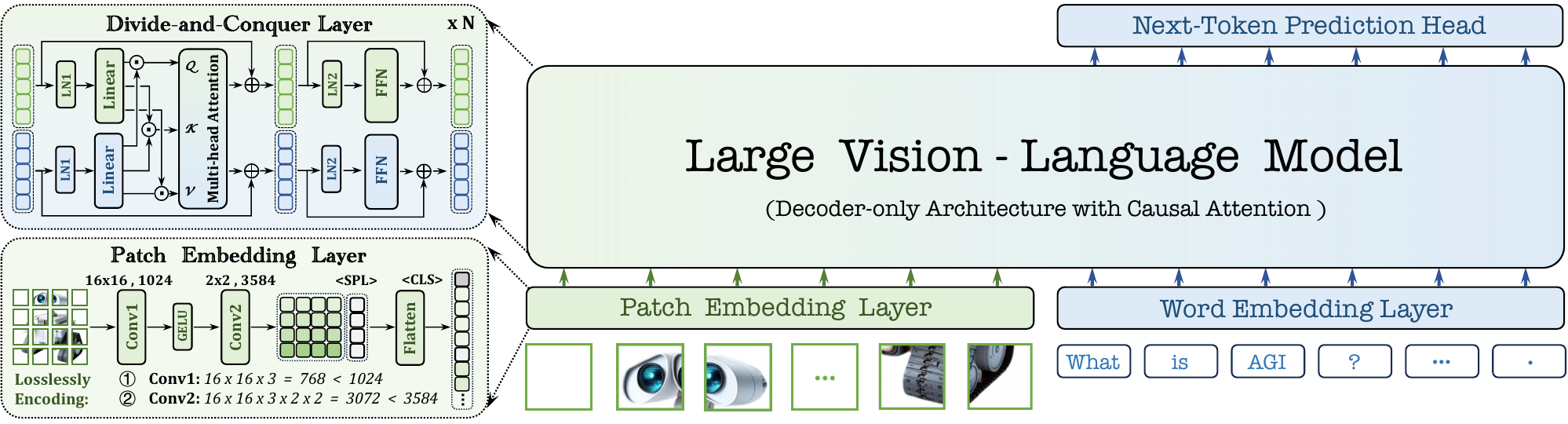} 
    \caption{Overview of our proposed EVEv2.0 framework. We first adopt a patch embedding layer to encode images losslessly, and then concatenate visual and textual tokens into a unified decoder-only vision-language model. Here, it extends the standard autoregressive transformer by incorporating modality-specific weights for each multi-head self-attention layer, feed-forward layer, and layer normalization.
    }
    \label{fig:framework}
\end{figure*}

\textbf{Finding (2):} \textit{Challenges towards multimodal interference and smooth transition.}
While mixing text-only and multi-modal data can alleviate this issue, it slows down the development of multi-modal understanding~\cite{VLM:DeepSeek-VL}.
Exp.(ii) shows that compared to EVEv1.0, EVEv1.2 approaches \textit{VE} in knowledge-related SQA-IMG metric, slightly mitigating catastrophic linguistic forgetting in LLMs. 
Hence, we explore potential architectures and training strategies targeting LLM adaptation and multimodal interference. EVEv1.2 with re-parameterization design for each feed-forward weight helps smooth the transition from LLMs to VLMs, while EVEv1.5 with MoE design helps decouple vision-language encoding heterogeneity.
However, after comparing LLMs and VLMs, we observe that \textit{VE} only requires minor adjustments to the LLM, whereas EVE necessitates extensive updates for similar capabilities in~\Cref{fig:preliminary}. 
Besides, Layer normalization emerges as the most impacted module, indicating that EVEv1.2 and EVEv1.5 struggle to efficiently construct visual representations and multi-modal alignments when attention and normalization layers in LLM are frozen. This dependency on LLM's pre-training paradigm limits the full potential for learning visual perception from scratch.

\subsection{Model Architecture}
\label{subsec:model_architecture}

Preliminary studies indicate that earlier EVE variants struggle to fully harness visual potential due to cross-modal interference within the pre-trained LLM distribution. To overcome this, we transform a dense transformer into a fully sparse, decoder-only architecture, guided by modality-aware routers in~\Cref{fig:framework}. This approach yields a heterogeneous, modality-mixing model that retains the computational structure and FLOP count of its dense transformer counterpart.

\textbf{Visual and Textual Encoding}.
For visual embeddings, we construct a minimalist patch embedding layer from scratch, eliminating strong inductive bias from pre-trained vision encoders in abstracting visual content.
Given an image input $I\in \mathbb{R}^{H\times W \times 3}$, we first employ a Convolution layer (Conv1), followed by a Gaussian Error Linear Unit (GELU) activation function. After obtaining the resulting 2-D feature map, we then adopt another Convolution layer (Conv2) to flexibly control computational complexity as follows: 
\begin{equation}
\label{eq:patch_embedding_layer}
x_{v}=\text{Conv2}(\text{GELU}(\text{Conv1}(I))),
\end{equation}
where $\text{Conv1}$ and $\text{Conv2}$ denote two convolutional layers with strides of 16 and 2, and output dimensions of 1024 and 3584, respectively. 
Thus, each pixel area (16×16×3×2×2 = 3072) is encoded into a 3584-dimensional feature, ensuring uncompressed and lossless information entropy.
Besides, large-kernel linear projections (\eg, 30×30) are hard to optimize, while 16×16 kernel quadruples visual tokens and increases overhead. Hence, two convolutional layers balance efficiency and accuracy.
Additionally, two special learnable tokens serve as the prompts for the image start and line feed. The class token $<\!\text{CLS}\!>$ is appended at the beginning of the image sequence, while the split tokens $<\!\text{SPL}\!>$ are inserted after each row of image tokens for indicators.
This patch embedding layer can support arbitrary-ratio images with up to about 2.5M pixels, \ie, 2.5K patch tokens. Afterward, we adopt the text tokenizer from Qwen2.5~\cite{qwen2.5} to encode text $T$ into token embeddings $x_{t}$ with a dimension of 3584.

\begin{figure*}[t]
    \centering 
    \includegraphics[width=0.98\linewidth,trim= 0 0 0 0,clip]
    {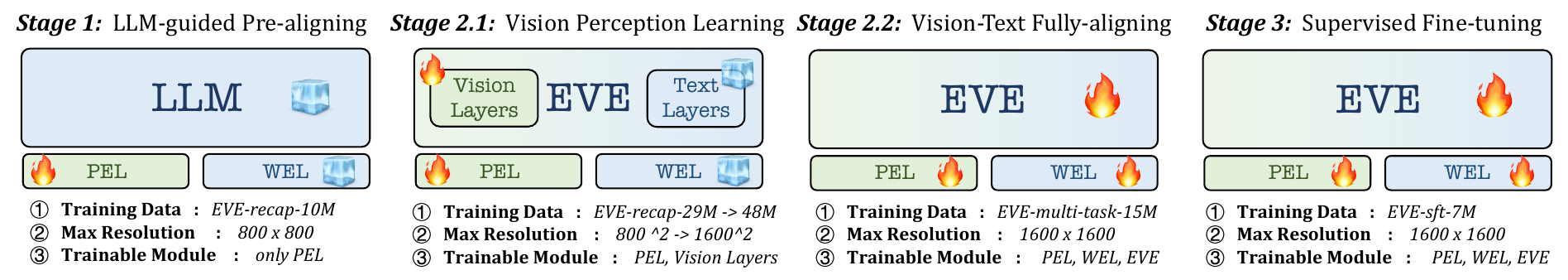} 
    \caption{Overview of training procedure. PEL/WEL denotes patch/word embedding layer. We begin by training the patch embedding layer to establish initial alignment across modalities. Afterward, we only update vision layers within the LLM to enhance visual perception progressively. Notably, we gradually increase the image resolutions from 800$\times$800 to 1600$\times$1600 and keep the original image aspect ratio. Finally, we train the entire model via QA and instruction data to strengthen cross-modality correspondence and complex understanding.}
    \label{fig:train_stage}
    \vspace{-2mm}
\end{figure*}

\textbf{Divide-and-Conquer Design.} 
Building on prior analyses, we propose explicitly decoupling key modules by introducing modality-aware components, including separate attention matrices (query, key, and value), normalization layers, and feed-forward modules, each with distinct parameters. 
Given the token sequence $x=(x_1, ..., x_n)$ where $x_i$ belongs to one specific modality $u_{i}\in \{v, t\}$, we perform a multi-head self-attention (ATTN) across all modalities, modeling cross-modal relationships in a unified feature space:
\begin{equation}
\begin{split}
\label{eq:divide_and_conquer}
\text{ATTN}(x;\{\theta_\text{attn}^u\}) &= \left(\text{softmax}\left(\frac{QK^T}{\sqrt{d_k}}\right)V\right)W_O^{u_{i}}, \\
Q_{i} = x_{i} W_Q^{u_{i}}&, K_{i} = x_{i} W_K^{u_{i}}, V_{i} = x_{i} W_V^{u_{i}},
\end{split}
\end{equation}
where modality-specific query, key, and value are derived from their respective attention weight matrices $W^{u_{i}}, u_{i}\in \{v, t\}$. 
The interaction process is performed across modalities, \ie, visual $x_v$ and textual $x_t$ sets. 
Inspired by~\Cref{fig:preliminary}, the significant quantified weight changes highlight the importance of decoupling the LayerNorm (LN) and Feed-Forward (FFN) layers; otherwise, they cause representational interference, limiting mutual capacities and capabilities. After fully decoupling the architecture, the overall operations within the Transformer block are defined as follows:
\begin{equation}
\begin{split}
    h&=x+\text{ATTN}(
    \text{LN1}(x;{\theta^u_{\text{ln1}}});
    \{\theta_\text{attn}^u\}),
    \\
    x^{\prime}&=h+\text{FFN}(
    \text{LN2}(x;{\theta^u_{\text{ln2}}});
    \theta_{\text{ffn}}^u).
\end{split}
\end{equation}
Compared with earlier EVE variants, EVEv2.0 employs a comprehensive decomposition with modality-specific components.
This minimizes interference in the representation space by fully unbinding each layer and processing token sets separately for each modality. The structural decomposition supports efficient vision-perception training from scratch while retaining pre-trained knowledge by freezing off-the-shelf LLMs during pretraining. This also allows independent single-modality encoding and cross-modality correspondence across different layers simultaneously, enabling flexible modeling patterns for understanding and reasoning.

\subsection{Training Procedure}
\label{subsec:training_procedure}

We divide the training process into four sequential stages in~\Cref{fig:train_stage}. The training data consists of publicly available image datasets, along with diverse question-answering (QA) datasets and multimodal dialogue data in~\Cref{tab:training_dataset}.

\textbf{DenseFusion++.} 
Developing strong visual perception requires high-quality, hyper-detailed image-text training data. Hence, we propose an efficient and low-budget engine based on LLaVA-1.6 (7B)~\cite{VLM:LLaVA-1.6} for scaling up synthetic data.
Specifically, it leverages multiple vision experts (\eg, Tag, Detection, OCR, ...) and learns the fusion strategy from GPT-4V, greatly facilitating data-scaling efficiency for native VLMs with high-quality annotations, not just distilling LLaVA.
We empirically validate the superior training efficiency achieved by using intensive annotations from DenseFusion++ during pre-training. It consistently outperforms original web-sourced captions or those generated by lower-quality caption engines. We will release the code and weight of the caption engine for better interpretation and further exploration.

\textbf{LLM-guided Pre-aligning.}
Following~\cite{VLM:EVE}, we freeze the LLM weights and train only the patch embedding layer to prevent model collapse and accelerate convergence in subsequent stages.
Using publicly available web data, we filter 44M image samples from Datacomp~\cite{Datasets:datacomp} recaptioned with our captioning engine. 
For training, we utilize a subset of 10M image-text pairs, dubbed \textit{EVE-recap-10M}, and optimize with cross-entropy (CE) loss. 
Our experiments suggest that more extensive training at this stage is beneficial for training stability, especially considering stronger LLMs.

\begin{table}[!t]
    \centering
    \caption{Details of training datasets across all stages. Note that we construct DenseFusion++ to re-caption web-scale image-text data.}
    \label{tab:training_dataset}
    \scalebox{1}{
    \begin{tabular}{c|c|c|c}
        \toprule
        Stage &Dataset &\#Num &Total\\
        \midrule
        \multirow{4}{*}{1 / 2.1} &Datacomp~\cite{Datasets:datacomp} &44M &\multirow{4}{*}{77M} \\
        &LAION~\cite{Datasets:Laion-5b} &15M &\\
        &SA-1B~\cite{TransF:SAM} &11M &\\
        &OpenImages~\cite{Datasets:OpenImages} &7M &\\
        \midrule
        2.2 &Infinity-MM-GeneralQA~\cite{gu2024infinity} &15M &15M\\
        \midrule
        \multirow{2}{*}{3} &LLaVA-onevision~\cite{Llava-onevision} &3.5M &\multirow{2}{*}{7.3M}\\
        &Infinity-MM-instruct~\cite{gu2024infinity} &3.8M &\\
        \bottomrule
    \end{tabular}
    }
\end{table}

\textbf{Vision Perception Learning.}
In this stage, we initialize the vision layers inside the LLM by loading LLM weights, where only the patch embedding and vision layers are trainable, while the Qwen2.5~\cite{qwen2.5} model remains frozen during training.
This strategy enables efficient learning of visual representations through pre-training on large-scale synthetic data, without compromising the knowledge encoded in the pre-trained LLM. Besides, we carefully partition the training data into a progressive, coarse-to-fine visual learning process. 
For training, we first introduce 29M re-captioning data from Datacomp~\cite{Datasets:datacomp} supervised by CE loss with a maximum of 640K image pixels (\ie, 625 patch tokens). Afterward, we increase the maximum image resolution to 2.5M pixels (\ie, 2.5K patch tokens) on an expanded dataset comprising 15M Datacomp~\cite{Datasets:datacomp}, 15M LAION~\cite{Datasets:Laion-5b}, 11M SA-1B~\cite{TransF:SAM}, and 7M OpenImages~\cite{Datasets:OpenImages}, dubbed \textit{EVE-recap-48M}.

\textbf{Vision-Text Fully-aligning.}
After establishing an initial alignment across modalities, we update the entire architecture to further improve image-text associations via the same loss functions. 
To facilitate this, we curate a diverse dataset of 15M samples from Infinity-MM general visual instruction~\cite{gu2024infinity}, including chart comprehension, OCR recognition, mathematical reasoning, and \etc. This dataset, named \textit{EVE-multi-task-15M}, enhances visual perception and vision-language alignment, equipping EVE with the foundational capabilities to handle various multimodal tasks.

\begin{table*}[th]
    \vspace{-2mm}
    \caption{Comparison with existing vision-language models on various vision-language benchmarks, including MMMU~\cite{Datasets:MMMU};
    MMB$^\text{en}$: MMBench-EN~\cite{Datasets:MMBench};
    SEED$^\text{I}$: SEEDBench-IMG~\cite{Datasets:Seed-bench};
    MMV: MMVet~\cite{Datasets:MM-vet};
    MME~\cite{Datasets:MME};
    POPE~\cite{Datasets:POPE};
    GQA~\cite{Datasets:GQA};  
    SQA$^\text{I}$: ScienceQA-IMG~\cite{Datasets:ScienceQA}; 
    TVQA: TextVQA~\cite{Datasets:TextVQA};   
    CQA: ChartQA~\cite{Datasets:ChartQA};
    AI2D~\cite{Datasets:AI2D};
    RWQA: RealWorldQA~\cite{Datasets:Realworldqa};
    OCRB: OCRBench~\cite{Datasets:OCRBench}.
    Note that \#A-Param denotes the number of activated parameters;
    \#Data represents the pre-training / fine-tuning data volume;
    \#Vtoken indicates the maximum image patch tokens.
    For MME, we sum up the perception and cognition scores.
    The best two results are marked in \textbf{bold} and \underline{underline}.
    }
    \label{tab:multimodal_benchmark}
    \vspace{-2mm}
    \centering
    \setlength{\tabcolsep}{0.075cm}
    \resizebox{\linewidth}{!}{
    \begin{tabular}{l ccc| cc ccccc cccc cc}
        \toprule
        Method & \#A-Param &\#Data &\#Vtoken
        & MMMU & MMB$^\text{en}$ 
        & SEED$^\text{I}$ & MMV & MME & POPE
        & GQA & SQA$^\text{I}$ & TQA & CQA 
        & AI2D & RWQA & OCRB \\
        \midrule
        \rowcolor{gray!17}
        \multicolumn{4}{l|}{\emph{Encoder-based Vision-Language Models:}} &\multicolumn{13}{l}{}\\
        InternVL-1.5
        & 2.2B &-- / -- &3328
        & 34.6 &70.9
        & 69.8 & 39.3 &\underline{1902} &\textbf{88.3}
        & 61.6 & \underline{84.9} &\underline{70.5} &\underline{74.8}
        &69.8 &57.9 & \underline{654}
        \\
        QwenVL-Chat
        & 7B &7.2B / 50M &256
        & 35.9 & 60.6 
        & 58.2 & -- &1848 &--
        & 57.5 & 68.2 & 61.5 & 49.8 
        & 45.9 & 49.3 & 488
        \\
        LLaVA-1.5
        & 7B &0.4B+ / 665K &576
        & 35.3 & 64.3 
        & 64.3 & 30.5 &1859 &85.9
        & 62.0 & 66.8 & 46.1 & 18.2 
        & 54.8 & 54.8 & 318
        \\
        LLaVA-1.6
        &7B &0.4B+ / 760K &2880
        & 35.1 & 67.4 
        & 64.7 & \underline{43.9} &1842 &\underline{86.4}
        & \underline{64.2} & 70.2 & 64.9 & 54.8 
        & 66.6 & 57.8 & 532
        \\
        Cambrian 
        &7B &10B+ / 7M &576
        &\underline{42.7} &\underline{75.9} 
        &\underline{74.7} &--  &-- &--
        &\textbf{64.6} &80.4 &\textbf{71.7} &73.3
        &\underline{73.0} &\underline{64.2} & 614
        \\
        LLaVA-OV 
        &7B &10B+ / 3.2M &7290
        &\textbf{47.3} &\textbf{81.7}  
        &\textbf{74.8} &\textbf{58.8} &\textbf{1998} &--
        &-- &\textbf{96.6} &-- &\textbf{78.8}
        &\textbf{81.6} &\textbf{65.5} & \textbf{697}
        \\
        \midrule
        \rowcolor{gray!17}
        \multicolumn{4}{l|}{\emph{Encoder-free Vision-Language Models:}}  &\multicolumn{13}{l}{}\\
        Fuyu 
        & 8B &-- / -- &--
        & 27.9  & 10.7 
        &  59.3   & 21.4 &-- & 84.0
        & --   &  56.8 & -- & --  
        & 64.5 & 43.7 & 366
        \\
        Chameleon
        &7B &1.4B+ / 1.8M &1024
        &25.4 &31.1   
        &30.6 & 8.3 &170 & 19.4
        & --  & 47.2 &4.8 &2.9
        &46.0 & 39.0 & 7.0 
        \\
        EVE
        &7B &33M / 1.8M &2304
        & 32.6 & 52.3
        & 64.6 & 25.7 & 1628 & 85.0  
        & \underline{62.6} &64.9 & 56.8 & 59.1  
        & 61.0 & -- & 398
        \\
        SOLO
        & 8B &43.7M / 2M &1024
        & --  &  \textbf{67.7}
        & 64.4 & 30.4 & 1260 &78.6
        & --   &73.3   & 25.0 & --  
        & 61.4 & 44.7 & 126 
        \\
        Mono-InternVL
        & 1.8B &1.3B / 7M &6400
        & \underline{33.7}  & 65.5
        & 67.4  & \underline{40.1} & \textbf{1875} &--
        & 59.5 &\underline{93.6}  &\textbf{72.6}  & \underline{73.7}  
        & 68.6 & -- & \textbf{767}
        \\
        Emu3
        & 8B &-- / -- &16K
        & 31.6 & 58.5
        & \underline{68.2} & 37.2 & -- & \underline{85.2}
        & 60.3 & 89.2 & 64.7 & 68.6 
        & \underline{70.0} & \underline{57.4} & 687 
        \\
        \textbf{EVEv2.0}
        & 7B &92M / 7.3M &2500
        &\textbf{39.3} &\underline{66.3} 
        &\textbf{71.4} &\textbf{45.0} & \underline{1709} & \textbf{87.6}
        &\textbf{62.9} &\textbf{96.2} &\underline{71.1} &\textbf{73.9}
        &\textbf{74.8} &\textbf{62.4} &\underline{702}
        \\
        \bottomrule
        \\
        \end{tabular}
        }
    \vspace{-2mm}
\end{table*}

\textbf{Supervised Fine-tuning.}
During the SFT stage, we further enhance EVE's ability to understand complex linguistic instructions and multifarious dialogue patterns, which are crucial for real-world applications. 
Here, we optimize the overall network layers on a diverse set of high-quality, multi-source instruction datasets, namely \textit{EVE-sft-7M}, including LLaVA-onevision~\cite{Llava-onevision} and partial Infinity-MM-instruct~\cite{gu2024infinity}. 
Note that Stages 2.2 and 3 can be merged if large, balanced, and high-quality instruction data is available. We separate them to handle diverse but uneven (Stage 2.2) and balanced (Stage 3) data to achieve consistent performance.

\section{Experiments}
\label{sec:Experiments}

\subsection{Training Settings}

\textbf{Data Preparation.} 
All the training data is collected from publicly accessible sources to ensure reproducibility.
\textit{(1) Image-Text Datasets.} We follow the pre-processing pipeline outlined in~\cite{VLM:EVE} to process SA-1B~\cite{TransF:SAM}, OpenImages~\cite{Datasets:OpenImages}, and LAION~\cite{Datasets:Laion-5b}, resulting in a total of about 33M samples. For Datacomp~\cite{Datasets:datacomp}, we curate the images with resolutions greater than $512 \times 512$, using DenseFusion++ to obtain 44M high-quality image descriptions and abandon samples with repetitive text or incomplete sentences.
\textit{(2) Question-answering and Instruction-following Datasets.} We clean out 15M QA data from Infinity-MM-GeneralQA~\cite{gu2024infinity} in its Stage-2. Meanwhile, we collect a blended set of the LLaVA-onevision~\cite{Llava-onevision} and partial Infinity-MM-instruct~\cite{gu2024infinity} from its original Stage-3/4 for complicated conversation patterns.

\textbf{Implementation Details.}
We use sixteen 8-A100 (40G) nodes to train EVEv2.0 using AdamW optimizer~\cite{Training:Adam}. For Stage 1, 2.1, 2.2, and 3, the batch sizes are 1024, 1024, 512, and 512, while the maximum learning rates are set to $2 \times 10^{-4}$, $1 \times 10^{-4}$, $2 \times 10^{-5}$, and $1 \times 10^{-5}$. We adopt warm-up strategy with the ratio of 0.03 and cosine decay scheduler across all stages. Unless otherwise stated, we set image resolutions as 800$^2$ and report fine-tuned results by LLaVA-mix-665K~\cite{VLM:LLaVA-1.5} for Stage 1/2.1/2.2 in~\Cref{sec:ablation_studies}.

\subsection{Main Results}
\label{subsec:main_results}

We conduct standard evaluations using the LMMs-Eval~\cite{zhang2024lmms-eval} across various vision-language benchmarks, including (1) Chart, Diagram, and Document Understanding tasks: OCRBench~\cite{Datasets:OCRBench}, ChartQA~\cite{Datasets:ChartQA}, and AI2D~\cite{Datasets:AI2D}; 
(2) Visual Perception and Challenging Reasoning tasks:
MMMU~\cite{Datasets:MMMU}, MMBench-EN~\cite{Datasets:MMBench}, SEEDBench-IMG~\cite{Datasets:Seed-bench}, MMVet~\cite{Datasets:MM-vet}, MME~\cite{Datasets:MME}, POPE~\cite{Datasets:POPE}, GQA~\cite{Datasets:GQA}, ScienceQA-IMG~\cite{Datasets:ScienceQA}, and TextVQA~\cite{Datasets:TextVQA}.
All the results are reported with greedy decoding and zero-shot settings, unless otherwise stated.

\begin{figure*}[t]
    \begin{minipage}[t]{0.44\textwidth}
    \centering 
    \includegraphics[width=\linewidth,trim= 0 0 0 0,clip]
    {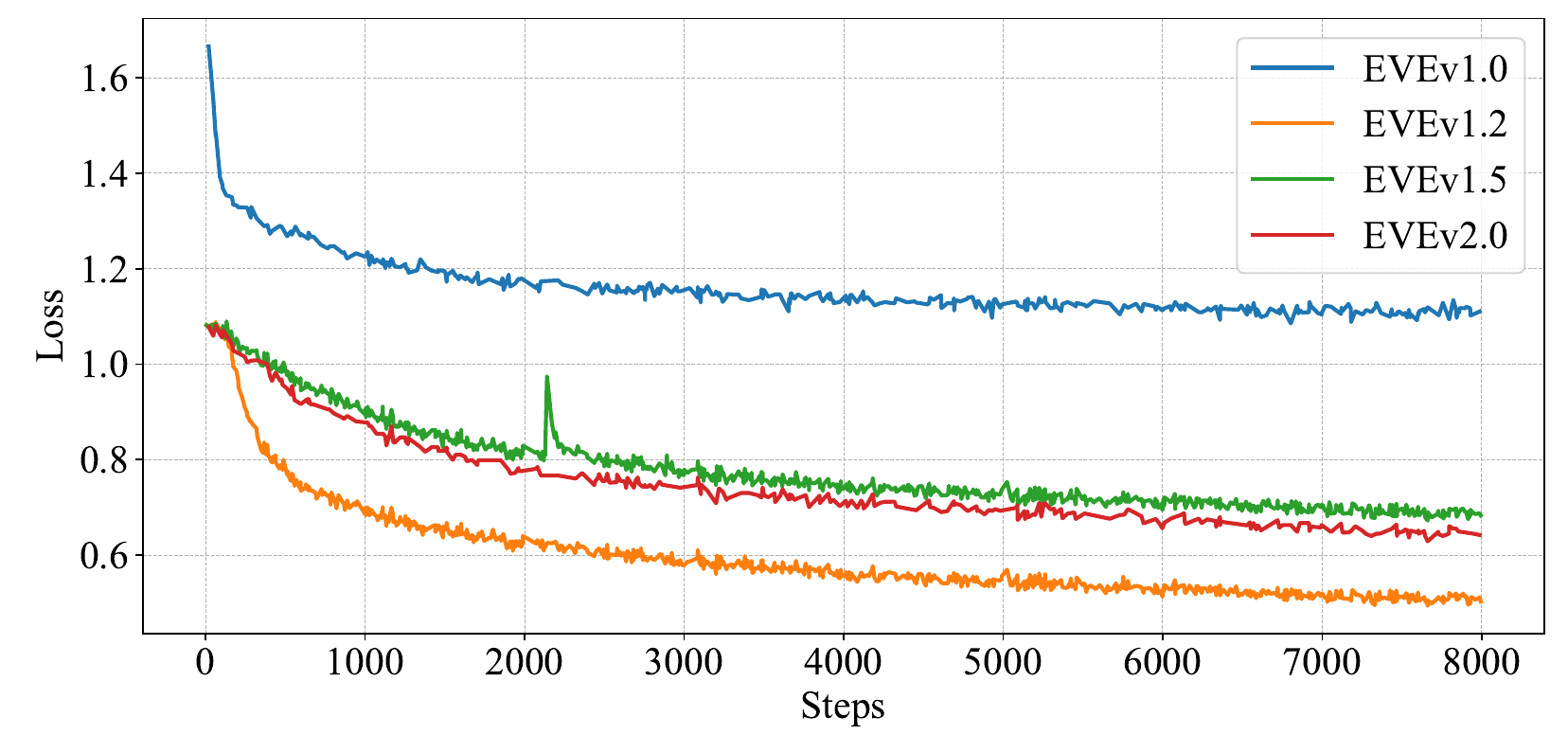} 
    \end{minipage}
    \hfill
    \begin{minipage}[t]{0.54\textwidth}
    \centering 
    \includegraphics[width=\linewidth,trim= 0 0 0 20,clip]
    {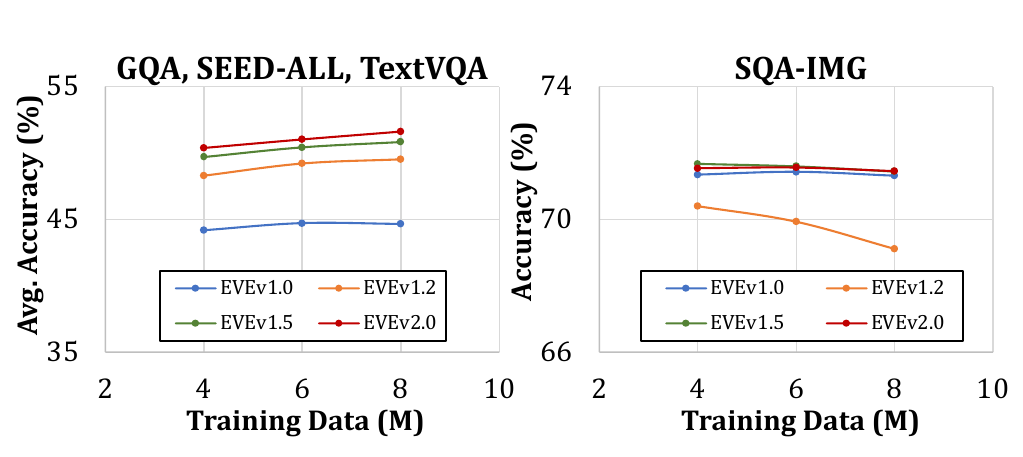} 
    \end{minipage}
    \caption{Training loss curve and evaluation results in Stage 2. We adopt various EVE variants based on Qwen-2.5~\cite{qwen2.5} as the baseline. We first train the patch embedding layer using \textit{EVE-recap-10M} in Stage 1, and further unfreeze vision layers except LLM layers in Stage 2.}
    \label{fig:different_version}
\end{figure*}

In~\Cref{tab:multimodal_benchmark}, EVEv2.0 surpasses encoder-free counterparts, \eg Fuyu~\cite{VLM:Fuyu-8b}, EVE~\cite{VLM:EVE}, SOLO~\cite{chen2024solo}, \etc across various vision-language benchmarks. 
Besides, Mono-InternVL~\cite{luo2024mono}, despite its smaller model size, achieves strong performance by leveraging 13x more pretraining data and large-scale, high-quality SFT data.
Notably, preliminary experiments in~\Cref{fig:preliminary} confirm that merely decoupling the feed-forward module is insufficient to address vision-language conflicts and module compatibility within a single unified network.
Furthermore, we clarify and validate that our Divide-and-Conquer architecture can achieve better data-scaling efficiency over the Mixture-of-Experts approach employed in Mono-InternVL~\cite{luo2024mono}, as demonstrated in~\Cref{fig:different_version}.

\begin{figure*}[t]
    \centering 
    \includegraphics[width=\linewidth,trim= 0 0 0 20,clip]
    {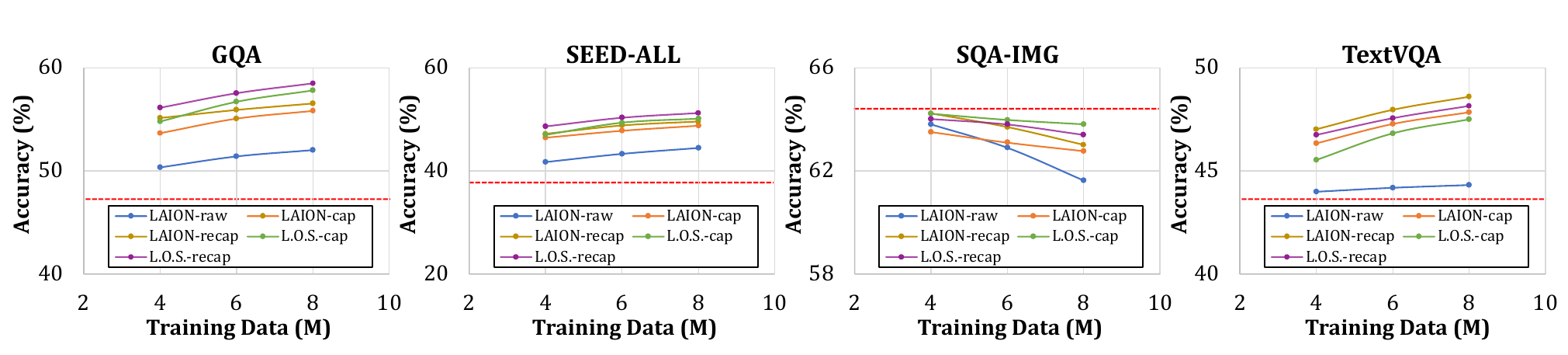} 
    \caption{Evaluation results of different data sources and caption engines. We utilize EVEv1.0 based on Vicuna-7B~\cite{TransF:Vicuna} as the baseline.
    Here ``*-raw'', ``*-cap'', or ``*-recap'' denote noisy web image captions, the samples annotated by both LLaVA-1.5 (13B) and Emu2 (17B), or modified DenseFusion++ (7B), respectively. Note that ``L.O.S.'' represents the mixture of LAION~\cite{Datasets:Laion-5b}, OpenImages~\cite{Datasets:OpenImages}, and SAM~\cite{TransF:SAM}.}
    \label{fig:data_source}
\end{figure*}

Besides, EVEv2.0 displays superior performance against the VLMs using discrete tokenizers, \ie Chameleon~\cite{team2024chameleon} and Emu3~\cite{VLM:EMU3}, despite being trained on significantly fewer data or utilizing fewer visual tokens. This further validates the efficiency and effectiveness of encoder-free VLMs with lossless visual encoding mode, even using a lightweight patch embedding layer from scratch. 
Notably, EVEv2.0 competes with popular and mainstream encoder-based VLMs, \eg LLaVA-1.6~\cite{VLM:LLaVA-1.6} and Cambrian~\cite{tong2024cambrian}. 
The performance gap with advanced modular VLMs stems largely from the significant discrepancy in in data scale. 
EVEv2.0 addresses this through modality-aware decomposition with high-quality annotations, offering an efficient and practical pathway for native VLMs to compete with encoder-based models.

\subsection{Ablation Studies}
\label{sec:ablation_studies}

\textbf{Divide-and-conquer (DaC) design outperforms the re-parameterization (ReP) and mixture-of-experts (MoE).} 
As shown in~\Cref{fig:different_version}, 
\textbf{(1)} the training process of EVEv1.0 is the slowest, and training only the patch embedding layer proves insufficient, resulting in minimal performance gains.
\textbf{(2)} EVEv1.2 (ReP) shows a rapid loss decrease, which can be attributed to the gradual transfer of LLMs into the initial VLMs by updating the pre-trained LLM weights. However, this approach leads to a noticeable performance drop on the SQA-IMG task requiring abundant text-related knowledge.
\textbf{(3)} In contrast, EVEv1.5 (MoE) only updates visual parameters inside frozen LLMs to effectively mitigate catastrophic forgetting issues during pre-training.
However, solely decoupling FFN modules restricts distinct feature distributions across modalities, resulting in less-efficient improvements in visual perception and multi-modality alignment.
\textbf{(4)} With prior validation support, EVEv2.0 (DaC) achieves optimal improvements across all multi-modal benchmarks, highlighting its superior data-scaling efficiency during large-scale pre-training. This success can be attributed to its modality-wise sparsity, which effectively preserves linguistic knowledge while providing greater flexibility for visual learning.
This philosophy is further evidenced by the loss curve in~\Cref{fig:different_version} with faster convergence and better training stability than EVEv1.5 during pre-training. 
\textbf{(5)} Notably, their Avg. accuracy gap rises from 0.8\% to 1.4\% as the training data grows from 8M to 24M, a trend likely to hold for other model sizes and data sources. Besides, only decoupling LayerNorm yields the Avg. accuracy of 48.8\% vs. 51.6\% for EVEv2.0 using 8M data, necessitating the complete decomposition.

\textbf{Fully-upgraded captioning engine facilitates training efficiency and model capabilities than prior competitors.} 
As illustrated in~\Cref{fig:data_source}, 
\textbf{(1)} web-scale image-text data often suffers from excessive noise and overly brief descriptions, which results in slow progress in visual content understanding and significantly pollutes LLM's pre-training knowledge. In contrast, using a powerful captioning engine to build high-quality, hyper-detailed image annotations proves essential for efficiently developing visual perception from scratch. Our modified DenseFusion++ (7B) outperforms previously adopted models like LLaVA-1.5 (13B) and Emu2 (17B) in this regard. Moreover, our model offers an additional advantage: its efficient and low-budget nature, capable of generating 700K descriptions per day with just a single 8-A100 (40G) node, accelerated by SGlang~\cite{VLM:SGlang}.
\textbf{(2)} A multi-source data mixture can significantly facilitate the visual training process. Our filtered LAION~\cite{Datasets:Laion-5b}, OpenImages~\cite{Datasets:OpenImages}, SAM~\cite{TransF:SAM} provide OCR-related images, real-world scenarios, and abundant image content, respectively. Together, these data mixtures can enhance the capability of VLMs to handle diverse image inputs, promoting the development of a more robust and versatile visual perception.

\begin{figure}[t]
    \centering 
    \includegraphics[width=0.9\linewidth,trim= 21 0 20 20,clip]
    {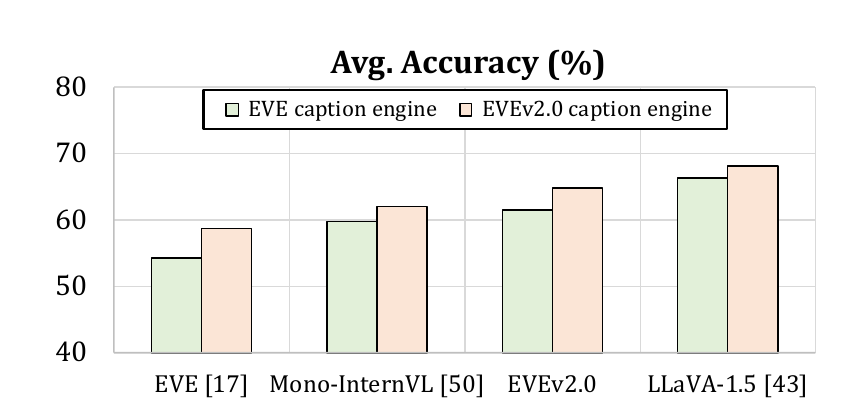} 
    \caption{Evaluation results of different methods. We report results on GQA, SEED-ALL, TextQA, SQA-IMG, MMB, and MMVet.}
    \label{fig:model_compare}
\end{figure}

\begin{figure}[t]
    \centering 
    \includegraphics[width=\linewidth,trim= 0 0 0 20,clip]
    {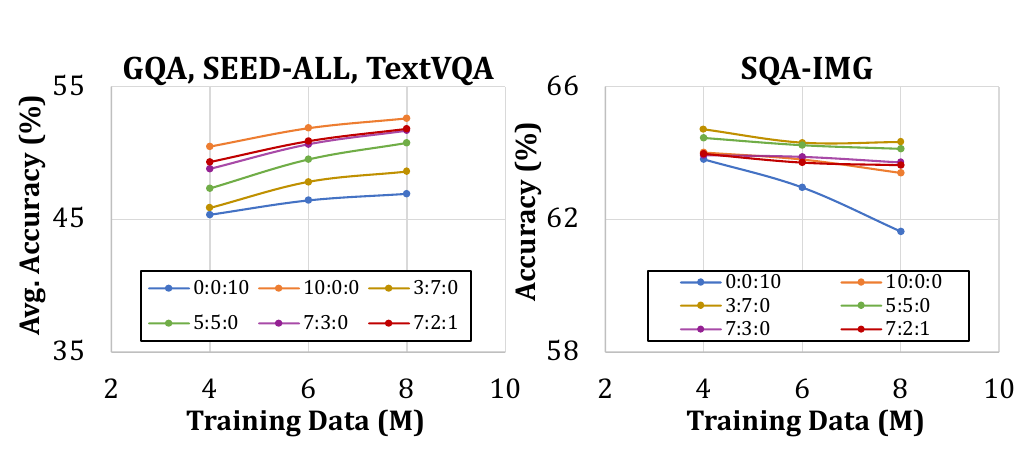} 
    \caption{Evaluation results of mixed data ratio. We adopt EVEv1.0 with Vicuna-7B~\cite{TransF:Vicuna} for validation. Note that x:y:z denote the proportion of synthesized data : language-only data : web data.}
    \label{fig:mixed_data_ratio}
\end{figure}

\textbf{EVEv2.0 achieves notable performance, rivaling the modular backbone equipped with strong visual encoders.}
In~\Cref{fig:model_compare}, we utilize Qwen-2.5~\cite{qwen2.5} as the LLM with \textit{EVE-recap-48M} for pre-training and \textit{EVE-sft-7M} for fine-tuning.
\textbf{(1)} EVEv2.0, with its modality-specific components and enhanced caption engine, effectively approaches the performance of encoder-based LLaVA-1.5~\cite{VLM:LLaVA-1.5}. It surpasses the prototype EVE~\cite{VLM:EVE}, which relies on visual supervision and a unified dense framework, as well as Mono-InternVL~\cite{luo2024mono}, which suffers from insufficient architecture decomposition.
\textbf{(2)} The DenseFusion++ delivers consistent and significant gains across various types of VLMs, leveraging diverse visual experts to boost generalist VLMs' perception and data-scaling efficiency for native VLM construction.
These findings further clarify the philosophy behind EVEv2.0.

\textbf{Meticulously adjusting data proportional distribution facilitates the balanced improvements across modalities.} 
We explore maintaining pre-trained LLM knowledge from a data mixture perspective in~\Cref{fig:mixed_data_ratio}. Striking the right balance is crucial for achieving robust multimodal capabilities without significantly sacrificing language performance. However, this balance is delicate and often influenced by various factors, \eg image resolution, dataset composition, text sources, and the type of language model. In this paper, we address multi-modality compatibility from the model structure perspective, leveraging overall multi-modal synthesized data during pre-training. Nonetheless, we believe that combining these two could yield even greater benefits.

\begin{figure}[t]
    \centering 
    \includegraphics[width=0.9\linewidth,trim= 21 0 20 20,clip]
    {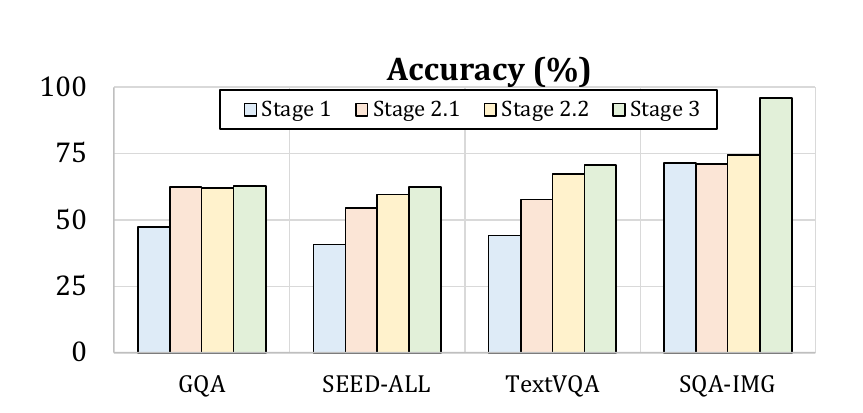} 
    \caption{Evaluation results across progressive training procedures. We adopt standard EVEv2.0 based on Qwen-2.5~\cite{qwen2.5} as the LLM.}
    \label{fig:training_process}
    \vspace{-9pt}
\end{figure}

\textbf{Incremental benefits across different training stages.} 
To thoroughly investigate progressive training recipes, we present the training dynamics with detailed recipes in~\Cref{tab:training_dataset}.
From~\Cref{fig:training_process}, we observe continual improvements with increasing pre-training data scales, which are further enhanced by carefully organized question-answering datasets.
After meticulously curating instruction-tuning datasets, our EVEv2.0 ultimately achieves superior multi-modality capabilities to handle various real-world application scenarios.

\section{Limitation and Discussion}
EVEv2.0 has systematically explored network architectures and training strategies for efficiently constructing encoder-free VLMs. Due to limitations in extensive high-quality data and computational devices, its full potential remains unrealized, thereby restricting performance on specific tasks, \eg knowledge- and document-oriented benchmarks. Besides, several promising directions remain for further exploration and improvement: Model Scaling, Data Scaling, and Modalities Expanding (\eg audio and video). We hope EVEv2.0 inspires further research on scaling laws for encoder-free VLMs with much more computational resources.

\section{Conclusion}
\label{sec:Conclusion}

In this paper, we propose EVEv2.0 with a transparent and powerful encoder-free architecture for vision-and-language understanding. Rather than focusing solely on state-of-the-art results, we systematically analyze the most efficient strategies for building visual perception from scratch.
To address vision-language interference, we fully disentangle the model components and introduce modality-wise sparsity within a unified decoder-only backbone.
Besides, we establish an efficient pathway for optimizing data-scaling efficiency in monolithic VLM research through a modified caption engine and a meticulous training recipe. 
Utilizing only 100M public data, EVEv2.0 outperforms existing encoder-free counterparts and steadily approaches mainstream encoder-based approaches of similar model capacity across various vision-language benchmarks. This provides valuable insights for developing scalable, native VLMs for the next generation.

\textbf{Acknowledgements.}
The paper is supported in part by the National Natural Science Foundation of China under grant No.62441231, 62293542, Liao Ning Province Science and Technology Plan No.2023JH26/10200016, Dalian City Science and Technology Innovation Fund No.2023JJ11CG001, and Natural Science Foundation of Zhejiang Province under Grant No.LD25F020001, Ningbo Key R\&D project under Grant No.2025Z039, and National Key R\&D Program of China (2022ZD0116302).
{
    \small
    \bibliographystyle{ieeenat_fullname}
    \bibliography{main}

\begin{thebibliography}{103}
\providecommand{\natexlab}[1]{#1}
\providecommand{\url}[1]{\texttt{#1}}
\expandafter\ifx\csname urlstyle\endcsname\relax
  \providecommand{\doi}[1]{doi: #1}\else
  \providecommand{\doi}{doi: \begingroup \urlstyle{rm}\Url}\fi

\bibitem[Alayrac et~al.(2022)Alayrac, Donahue, Luc, Miech, Barr, Hasson, Lenc, Mensch, Millican, Reynolds, Ring, Rutherford, Cabi, Han, Gong, Samangooei, Monteiro, Menick, Borgeaud, Brock, Nematzadeh, Sharifzadeh, Binkowski, Barreira, Vinyals, Zisserman, and Simonyan]{VLP:Flamingo}
Jean{-}Baptiste Alayrac, Jeff Donahue, Pauline Luc, Antoine Miech, Iain Barr, Yana Hasson, Karel Lenc, Arthur Mensch, Katherine Millican, Malcolm Reynolds, Roman Ring, Eliza Rutherford, Serkan Cabi, Tengda Han, Zhitao Gong, Sina Samangooei, Marianne Monteiro, Jacob~L. Menick, Sebastian Borgeaud, Andy Brock, Aida Nematzadeh, Sahand Sharifzadeh, Mikolaj Binkowski, Ricardo Barreira, Oriol Vinyals, Andrew Zisserman, and Kar{\'{e}}n Simonyan.
\newblock Flamingo: a visual language model for few-shot learning.
\newblock In \emph{NeurIPS}, 2022.

\bibitem[Anthropic(2024)]{VLM:Claude3}
AI Anthropic.
\newblock The claude 3 model family: Opus, sonnet, haiku.
\newblock \emph{Claude-3 Model Card}, 2024.

\bibitem[Bai et~al.(2023{\natexlab{a}})Bai, Bai, Chu, Cui, Dang, Deng, Fan, Ge, Han, Huang, Hui, Ji, Li, Lin, Lin, Liu, Liu, Lu, Lu, Ma, Men, Ren, Ren, Tan, Tan, Tu, Wang, Wang, Wang, Wu, Xu, Xu, Yang, Yang, Yang, Yang, Yao, Yu, Yuan, Yuan, Zhang, Zhang, Zhang, Zhang, Zhou, Zhou, Zhou, and Zhu]{TransF:Qwen}
Jinze Bai, Shuai Bai, Yunfei Chu, Zeyu Cui, Kai Dang, Xiaodong Deng, Yang Fan, Wenbin Ge, Yu Han, Fei Huang, Binyuan Hui, Luo Ji, Mei Li, Junyang Lin, Runji Lin, Dayiheng Liu, Gao Liu, Chengqiang Lu, Keming Lu, Jianxin Ma, Rui Men, Xingzhang Ren, Xuancheng Ren, Chuanqi Tan, Sinan Tan, Jianhong Tu, Peng Wang, Shijie Wang, Wei Wang, Shengguang Wu, Benfeng Xu, Jin Xu, An Yang, Hao Yang, Jian Yang, Shusheng Yang, Yang Yao, Bowen Yu, Hongyi Yuan, Zheng Yuan, Jianwei Zhang, Xingxuan Zhang, Yichang Zhang, Zhenru Zhang, Chang Zhou, Jingren Zhou, Xiaohuan Zhou, and Tianhang Zhu.
\newblock Qwen technical report.
\newblock \emph{arXiv: 2309.16609}, 2023{\natexlab{a}}.

\bibitem[Bai et~al.(2023{\natexlab{b}})Bai, Bai, Yang, Wang, Tan, Wang, Lin, Zhou, and Zhou]{VLM:qwen-vl}
Jinze Bai, Shuai Bai, Shusheng Yang, Shijie Wang, Sinan Tan, Peng Wang, Junyang Lin, Chang Zhou, and Jingren Zhou.
\newblock Qwen-vl: A versatile vision-language model for understanding, localization, text reading, and beyond.
\newblock \emph{arXiv preprint arXiv:2308.12966}, 1\penalty0 (2):\penalty0 3, 2023{\natexlab{b}}.

\bibitem[Bao et~al.(2022)Bao, Wang, Dong, Liu, Mohammed, Aggarwal, Som, Piao, and Wei]{bao2022vlmo}
Hangbo Bao, Wenhui Wang, Li Dong, Qiang Liu, Owais~Khan Mohammed, Kriti Aggarwal, Subhojit Som, Songhao Piao, and Furu Wei.
\newblock Vlmo: Unified vision-language pre-training with mixture-of-modality-experts.
\newblock \emph{Advances in Neural Information Processing Systems}, 35:\penalty0 32897--32912, 2022.

\bibitem[Bavishi et~al.(2023)Bavishi, Elsen, Hawthorne, Nye, Odena, Somani, and Ta\c{s}\i{}rlar]{VLM:Fuyu-8b}
Rohan Bavishi, Erich Elsen, Curtis Hawthorne, Maxwell Nye, Augustus Odena, Arushi Somani, and Sa\u{g}nak Ta\c{s}\i{}rlar.
\newblock Introducing our multimodal models, 2023.

\bibitem[Beyer et~al.(2024)Beyer, Steiner, Pinto, Kolesnikov, Wang, Salz, Neumann, Alabdulmohsin, Tschannen, Bugliarello, et~al.]{beyer2024paligemma}
Lucas Beyer, Andreas Steiner, Andr{\'e}~Susano Pinto, Alexander Kolesnikov, Xiao Wang, Daniel Salz, Maxim Neumann, Ibrahim Alabdulmohsin, Michael Tschannen, Emanuele Bugliarello, et~al.
\newblock Paligemma: A versatile 3b vlm for transfer.
\newblock \emph{arXiv preprint arXiv:2407.07726}, 2024.

\bibitem[Bi et~al.(2024)Bi, Chen, Chen, Chen, Dai, Deng, Ding, Dong, Du, Fu, et~al.]{TransF:Deepseekllm}
Xiao Bi, Deli Chen, Guanting Chen, Shanhuang Chen, Damai Dai, Chengqi Deng, Honghui Ding, Kai Dong, Qiushi Du, Zhe Fu, et~al.
\newblock Deepseek llm: Scaling open-source language models with longtermism.
\newblock \emph{arXiv: 2401.02954}, 2024.

\bibitem[Cai et~al.(2024)Cai, Cao, Chen, Chen, Chen, Chen, Chen, Chen, Chen, Chu, Dong, Duan, Fan, Fei, Gao, Ge, Gu, Gu, Gui, Guo, Guo, He, Hu, Huang, Jiang, Jiao, Jin, Lei, Li, Li, Li, Li, Li, Li, Liu, Liu, Hong, Liu, Liu, Liu, Lv, Lv, Lv, Ma, Ma, Ma, Ning, Ouyang, Qiu, Qu, Shang, Shao, Song, Song, Sui, Sun, Sun, Tang, Wang, Wang, Wang, Wang, Wang, Wang, Wang, Wei, Weng, Wu, Xiong, and et~al.]{TransF:InternLM2}
Zheng Cai, Maosong Cao, Haojiong Chen, Kai Chen, Keyu Chen, Xin Chen, Xun Chen, Zehui Chen, Zhi Chen, Pei Chu, Xiaoyi Dong, Haodong Duan, Qi Fan, Zhaoye Fei, Yang Gao, Jiaye Ge, Chenya Gu, Yuzhe Gu, Tao Gui, Aijia Guo, Qipeng Guo, Conghui He, Yingfan Hu, Ting Huang, Tao Jiang, Penglong Jiao, Zhenjiang Jin, Zhikai Lei, Jiaxing Li, Jingwen Li, Linyang Li, Shuaibin Li, Wei Li, Yining Li, Hongwei Liu, Jiangning Liu, Jiawei Hong, Kaiwen Liu, Kuikun Liu, Xiaoran Liu, Chengqi Lv, Haijun Lv, Kai Lv, Li Ma, Runyuan Ma, Zerun Ma, Wenchang Ning, Linke Ouyang, Jiantao Qiu, Yuan Qu, Fukai Shang, Yunfan Shao, Demin Song, Zifan Song, Zhihao Sui, Peng Sun, Yu Sun, Huanze Tang, Bin Wang, Guoteng Wang, Jiaqi Wang, Jiayu Wang, Rui Wang, Yudong Wang, Ziyi Wang, Xingjian Wei, Qizhen Weng, Fan Wu, Yingtong Xiong, and et al.
\newblock Internlm2 technical report.
\newblock \emph{arXiv: 2403.17297}, 2024.

\bibitem[Caron et~al.(2021)Caron, Touvron, Misra, J{\'{e}}gou, Mairal, Bojanowski, and Joulin]{CL:DINO}
Mathilde Caron, Hugo Touvron, Ishan Misra, Herv{\'{e}} J{\'{e}}gou, Julien Mairal, Piotr Bojanowski, and Armand Joulin.
\newblock Emerging properties in self-supervised vision transformers.
\newblock In \emph{ICCV}, pages 9630--9640, 2021.

\bibitem[Chen et~al.(2024{\natexlab{a}})Chen, Chen, Zhang, Chen, Wu, Zhang, Chen, Li, Wan, and Wang]{chen2024allava}
Guiming~Hardy Chen, Shunian Chen, Ruifei Zhang, Junying Chen, Xiangbo Wu, Zhiyi Zhang, Zhihong Chen, Jianquan Li, Xiang Wan, and Benyou Wang.
\newblock Allava: Harnessing gpt4v-synthesized data for a lite vision-language model.
\newblock \emph{arXiv preprint arXiv:2402.11684}, 2024{\natexlab{a}}.

\bibitem[Chen et~al.(2023{\natexlab{a}})Chen, Li, Dong, Zhang, He, Wang, Zhao, and Lin]{VLM:Sharegpt4v}
Lin Chen, Jisong Li, Xiaoyi Dong, Pan Zhang, Conghui He, Jiaqi Wang, Feng Zhao, and Dahua Lin.
\newblock Sharegpt4v: Improving large multi-modal models with better captions.
\newblock \emph{arXiv: 2311.12793}, 2023{\natexlab{a}}.

\bibitem[Chen et~al.(2024{\natexlab{b}})Chen, Wang, Peng, and Ji]{chen2024solo}
Yangyi Chen, Xingyao Wang, Hao Peng, and Heng Ji.
\newblock A single transformer for scalable vision-language modeling.
\newblock \emph{arXiv preprint arXiv:2407.06438}, 2024{\natexlab{b}}.

\bibitem[Chen et~al.(2023{\natexlab{b}})Chen, Wu, Wang, Su, Chen, Xing, Zhong, Zhang, Zhu, Lu, Li, Luo, Lu, Qiao, and Dai]{VLM:InternVL}
Zhe Chen, Jiannan Wu, Wenhai Wang, Weijie Su, Guo Chen, Sen Xing, Muyan Zhong, Qinglong Zhang, Xizhou Zhu, Lewei Lu, Bin Li, Ping Luo, Tong Lu, Yu Qiao, and Jifeng Dai.
\newblock Internvl: Scaling up vision foundation models and aligning for generic visual-linguistic tasks.
\newblock \emph{arXiv: 2312.14238}, 2023{\natexlab{b}}.

\bibitem[Chen et~al.(2024{\natexlab{c}})Chen, Wang, Tian, Ye, Gao, Cui, Tong, Hu, Luo, Ma, et~al.]{VLM:InternVL-1.5}
Zhe Chen, Weiyun Wang, Hao Tian, Shenglong Ye, Zhangwei Gao, Erfei Cui, Wenwen Tong, Kongzhi Hu, Jiapeng Luo, Zheng Ma, et~al.
\newblock How far are we to gpt-4v? closing the gap to commercial multimodal models with open-source suites.
\newblock \emph{arXiv:2404.16821}, 2024{\natexlab{c}}.

\bibitem[Chiang et~al.(2023)Chiang, Li, Lin, Sheng, Wu, Zhang, Zheng, Zhuang, Zhuang, Gonzalez, Stoica, and Xing]{TransF:Vicuna}
Wei-Lin Chiang, Zhuohan Li, Zi Lin, Ying Sheng, Zhanghao Wu, Hao Zhang, Lianmin Zheng, Siyuan Zhuang, Yonghao Zhuang, Joseph~E. Gonzalez, Ion Stoica, and Eric~P. Xing.
\newblock Vicuna: An open-source chatbot impressing gpt-4 with 90\%* chatgpt quality, 2023.

\bibitem[Dai et~al.(2023)Dai, Li, Li, Tiong, Zhao, Wang, Li, Fung, and Hoi]{VLM:InstructBLIP}
Wenliang Dai, Junnan Li, Dongxu Li, Anthony Meng~Huat Tiong, Junqi Zhao, Weisheng Wang, Boyang Li, Pascale Fung, and Steven C.~H. Hoi.
\newblock Instructblip: Towards general-purpose vision-language models with instruction tuning.
\newblock In \emph{NeurIPS}, 2023.

\bibitem[Dehghani et~al.(2023)Dehghani, Djolonga, Mustafa, Padlewski, Heek, Gilmer, Steiner, Caron, Geirhos, Alabdulmohsin, Jenatton, Beyer, Tschannen, Arnab, Wang, Ruiz, Minderer, Puigcerver, Evci, Kumar, van Steenkiste, Elsayed, Mahendran, Yu, Oliver, Huot, Bastings, Collier, Gritsenko, Birodkar, Vasconcelos, Tay, Mensink, Kolesnikov, Pavetic, Tran, Kipf, Lucic, Zhai, Keysers, Harmsen, and Houlsby]{TransF:ViT-22B}
Mostafa Dehghani, Josip Djolonga, Basil Mustafa, Piotr Padlewski, Jonathan Heek, Justin Gilmer, Andreas~Peter Steiner, Mathilde Caron, Robert Geirhos, Ibrahim Alabdulmohsin, Rodolphe Jenatton, Lucas Beyer, Michael Tschannen, Anurag Arnab, Xiao Wang, Carlos~Riquelme Ruiz, Matthias Minderer, Joan Puigcerver, Utku Evci, Manoj Kumar, Sjoerd van Steenkiste, Gamaleldin~Fathy Elsayed, Aravindh Mahendran, Fisher Yu, Avital Oliver, Fantine Huot, Jasmijn Bastings, Mark Collier, Alexey~A. Gritsenko, Vighnesh Birodkar, Cristina~Nader Vasconcelos, Yi Tay, Thomas Mensink, Alexander Kolesnikov, Filip Pavetic, Dustin Tran, Thomas Kipf, Mario Lucic, Xiaohua Zhai, Daniel Keysers, Jeremiah~J. Harmsen, and Neil Houlsby.
\newblock Scaling vision transformers to 22 billion parameters.
\newblock In \emph{ICML}, pages 7480--7512, 2023.

\bibitem[Diao et~al.(2024{\natexlab{a}})Diao, Cui, Li, Wang, Lu, and Wang]{VLM:EVE}
Haiwen Diao, Yufeng Cui, Xiaotong Li, Yueze Wang, Huchuan Lu, and Xinlong Wang.
\newblock Unveiling encoder-free vision-language models.
\newblock \emph{arXiv preprint arXiv:2406.11832}, 2024{\natexlab{a}}.

\bibitem[Diao et~al.(2024{\natexlab{b}})Diao, Wan, Zhang, Jia, Lu, and Chen]{TL:UniPT}
Haiwen Diao, Bo Wan, Ying Zhang, Xu Jia, Huchuan Lu, and Long Chen.
\newblock Unipt: Universal parallel tuning for transfer learning with efficient parameter and memory.
\newblock In \emph{CVPR}, 2024{\natexlab{b}}.

\bibitem[Diao et~al.(2025)Diao, Wan, Jia, Zhuge, Zhang, Lu, and Chen]{TL:SHERL}
Haiwen Diao, Bo Wan, Xu Jia, Yunzhi Zhuge, Ying Zhang, Huchuan Lu, and Long Chen.
\newblock Sherl: Synthesizing high accuracy and efficient memory for resource-limited transfer learning.
\newblock In \emph{ECCV}, pages 75--95, 2025.

\bibitem[Dosovitskiy et~al.(2021)Dosovitskiy, Beyer, Kolesnikov, Weissenborn, Zhai, Unterthiner, Dehghani, Minderer, Heigold, Gelly, Uszkoreit, and Houlsby]{TransF:ViT}
Alexey Dosovitskiy, Lucas Beyer, Alexander Kolesnikov, Dirk Weissenborn, Xiaohua Zhai, Thomas Unterthiner, Mostafa Dehghani, Matthias Minderer, Georg Heigold, Sylvain Gelly, Jakob Uszkoreit, and Neil Houlsby.
\newblock An image is worth 16x16 words: Transformers for image recognition at scale.
\newblock In \emph{ICLR}, 2021.

\bibitem[Esser et~al.(2021)Esser, Rombach, and Ommer]{VQGAN}
Patrick Esser, Robin Rombach, and Bjorn Ommer.
\newblock Taming transformers for high-resolution image synthesis.
\newblock In \emph{CVPR}, pages 12873--12883, 2021.

\bibitem[Fang et~al.(2023)Fang, Wang, Xie, Sun, Wu, Wang, Huang, Wang, and Cao]{TransF:EVA}
Yuxin Fang, Wen Wang, Binhui Xie, Quan Sun, Ledell Wu, Xinggang Wang, Tiejun Huang, Xinlong Wang, and Yue Cao.
\newblock {EVA:} exploring the limits of masked visual representation learning at scale.
\newblock In \emph{CVPR}, pages 19358--19369, 2023.

\bibitem[Fu et~al.(2023)Fu, Chen, Shen, Qin, Zhang, Lin, Qiu, Lin, Yang, Zheng, Li, Sun, and Ji]{Datasets:MME}
Chaoyou Fu, Peixian Chen, Yunhang Shen, Yulei Qin, Mengdan Zhang, Xu Lin, Zhenyu Qiu, Wei Lin, Jinrui Yang, Xiawu Zheng, Ke Li, Xing Sun, and Rongrong Ji.
\newblock {MME:} {A} comprehensive evaluation benchmark for multimodal large language models.
\newblock \emph{arXiv: 2306.13394}, 2023.

\bibitem[Gadre et~al.(2024)Gadre, Ilharco, Fang, Hayase, Smyrnis, Nguyen, Marten, Wortsman, Ghosh, Zhang, et~al.]{Datasets:datacomp}
Samir~Yitzhak Gadre, Gabriel Ilharco, Alex Fang, Jonathan Hayase, Georgios Smyrnis, Thao Nguyen, Ryan Marten, Mitchell Wortsman, Dhruba Ghosh, Jieyu Zhang, et~al.
\newblock Datacomp: In search of the next generation of multimodal datasets.
\newblock \emph{NeurIPS}, 36, 2024.

\bibitem[Gao et~al.(2023)Gao, Han, Zhang, Lin, Geng, Zhou, Zhang, Lu, He, Yue, et~al.]{VLM:Llama-Adapterv2}
Peng Gao, Jiaming Han, Renrui Zhang, Ziyi Lin, Shijie Geng, Aojun Zhou, Wei Zhang, Pan Lu, Conghui He, Xiangyu Yue, et~al.
\newblock Llama-adapter v2: Parameter-efficient visual instruction model.
\newblock \emph{arXiv: 2304.15010}, 2023.

\bibitem[Ge et~al.(2024)Ge, Zhao, Zeng, Ge, Li, Wang, and Shan]{VLM:SEED}
Yuying Ge, Sijie Zhao, Ziyun Zeng, Yixiao Ge, Chen Li, Xintao Wang, and Ying Shan.
\newblock Making {LL}a{MA} {SEE} and draw with {SEED} tokenizer.
\newblock In \emph{The Twelfth International Conference on Learning Representations}, 2024.

\bibitem[Gu et~al.(2024)Gu, Zhang, Zhou, Yu, Xing, Wang, Cao, Jia, Zhang, Wang, et~al.]{gu2024infinity}
Shuhao Gu, Jialing Zhang, Siyuan Zhou, Kevin Yu, Zhaohu Xing, Liangdong Wang, Zhou Cao, Jintao Jia, Zhuoyi Zhang, Yixuan Wang, et~al.
\newblock Infinity-mm: Scaling multimodal performance with large-scale and high-quality instruction data.
\newblock \emph{arXiv preprint arXiv:2410.18558}, 2024.

\bibitem[Hong et~al.(2023)Hong, Wang, Lv, Xu, Yu, Ji, Wang, Wang, Dong, Ding, et~al.]{VLM:CogAgent}
Wenyi Hong, Weihan Wang, Qingsong Lv, Jiazheng Xu, Wenmeng Yu, Junhui Ji, Yan Wang, Zihan Wang, Yuxiao Dong, Ming Ding, et~al.
\newblock Cogagent: A visual language model for gui agents.
\newblock \emph{arXiv:2312.08914}, 2023.

\bibitem[Hu et~al.(2024)Hu, Xu, Zhang, Ye, Yan, Zhang, Jin, Huang, and Zhou]{hu2024mplug-docowl2}
Anwen Hu, Haiyang Xu, Liang Zhang, Jiabo Ye, Ming Yan, Ji Zhang, Qin Jin, Fei Huang, and Jingren Zhou.
\newblock mplug-docowl2: High-resolution compressing for ocr-free multi-page document understanding.
\newblock \emph{arXiv preprint arXiv:2409.03420}, 2024.

\bibitem[Hudson and Manning(2019)]{Datasets:GQA}
Drew~A. Hudson and Christopher~D. Manning.
\newblock {GQA:} {A} new dataset for real-world visual reasoning and compositional question answering.
\newblock In \emph{CVPR}, pages 6700--6709, 2019.

\bibitem[{IDEFICS Research Team}(2023)]{VLM:IDEFICS}
{IDEFICS Research Team}.
\newblock Introducing idefics: An open reproduction of state-of-the-art visual language model.
\newblock \url{https://huggingface.co/blog/idefics}, 2023.

\bibitem[Kembhavi et~al.(2016)Kembhavi, Salvato, Kolve, Seo, Hajishirzi, and Farhadi]{Datasets:AI2D}
Aniruddha Kembhavi, Mike Salvato, Eric Kolve, Minjoon Seo, Hannaneh Hajishirzi, and Ali Farhadi.
\newblock A diagram is worth a dozen images.
\newblock In \emph{ECCV}, pages 235--251, 2016.

\bibitem[Kingma and Ba(2015)]{Training:Adam}
Diederik~P. Kingma and Jimmy Ba.
\newblock Adam: {A} method for stochastic optimization.
\newblock In \emph{ICLR}, 2015.

\bibitem[Kirillov et~al.(2023)Kirillov, Mintun, Ravi, Mao, Rolland, Gustafson, Xiao, Whitehead, Berg, Lo, Doll{\'{a}}r, and Girshick]{TransF:SAM}
Alexander Kirillov, Eric Mintun, Nikhila Ravi, Hanzi Mao, Chlo{\'{e}} Rolland, Laura Gustafson, Tete Xiao, Spencer Whitehead, Alexander~C. Berg, Wan{-}Yen Lo, Piotr Doll{\'{a}}r, and Ross~B. Girshick.
\newblock Segment anything.
\newblock \emph{arXiv: 2304.02643}, 2023.

\bibitem[Kuznetsova et~al.(2018)Kuznetsova, Rom, Alldrin, Uijlings, Krasin, Pont{-}Tuset, Kamali, Popov, Malloci, Duerig, and Ferrari]{Datasets:OpenImages}
Alina Kuznetsova, Hassan Rom, Neil Alldrin, Jasper R.~R. Uijlings, Ivan Krasin, Jordi Pont{-}Tuset, Shahab Kamali, Stefan Popov, Matteo Malloci, Tom Duerig, and Vittorio Ferrari.
\newblock The open images dataset {V4:} unified image classification, object detection, and visual relationship detection at scale.
\newblock \emph{arXiv: 1811.00982}, 2018.

\bibitem[Lauren{\c{c}}on et~al.(2024)Lauren{\c{c}}on, Marafioti, Sanh, and Tronchon]{laurenccon2024docmatix}
Hugo Lauren{\c{c}}on, Andr{\'e}s Marafioti, Victor Sanh, and L{\'e}o Tronchon.
\newblock Building and better understanding vision-language models: insights and future directions.
\newblock \emph{arXiv preprint arXiv:2408.12637}, 2024.

\bibitem[Li et~al.(2023{\natexlab{a}})Li, Wang, Wang, Ge, Ge, and Shan]{Datasets:Seed-bench}
Bohao Li, Rui Wang, Guangzhi Wang, Yuying Ge, Yixiao Ge, and Ying Shan.
\newblock Seed-bench: Benchmarking multimodal llms with generative comprehension.
\newblock \emph{arXiv: 2307.16125}, 2023{\natexlab{a}}.

\bibitem[Li et~al.(2024{\natexlab{a}})Li, Zhang, Guo, Zhang, Li, Zhang, Zhang, Li, Liu, and Li]{Llava-onevision}
Bo Li, Yuanhan Zhang, Dong Guo, Renrui Zhang, Feng Li, Hao Zhang, Kaichen Zhang, Yanwei Li, Ziwei Liu, and Chunyuan Li.
\newblock Llava-onevision: Easy visual task transfer.
\newblock \emph{arXiv preprint arXiv:2408.03326}, 2024{\natexlab{a}}.

\bibitem[Li et~al.(2024{\natexlab{b}})Li, Liu, Wu, Wang, Shen, Qu, Niu, Wang, Chen, and Li]{li2024aria}
Dongxu Li, Yudong Liu, Haoning Wu, Yue Wang, Zhiqi Shen, Bowen Qu, Xinyao Niu, Guoyin Wang, Bei Chen, and Junnan Li.
\newblock Aria: An open multimodal native mixture-of-experts model.
\newblock \emph{arXiv preprint arXiv:2410.05993}, 2024{\natexlab{b}}.

\bibitem[Li et~al.(2022{\natexlab{a}})Li, Li, Xiong, and Hoi]{VLP:BLIP}
Junnan Li, Dongxu Li, Caiming Xiong, and Steven C.~H. Hoi.
\newblock {BLIP:} bootstrapping language-image pre-training for unified vision-language understanding and generation.
\newblock In \emph{ICLR}, pages 12888--12900, 2022{\natexlab{a}}.

\bibitem[Li et~al.(2023{\natexlab{b}})Li, Li, Savarese, and Hoi]{VLP:BLIPv2}
Junnan Li, Dongxu Li, Silvio Savarese, and Steven C.~H. Hoi.
\newblock {BLIP-2:} bootstrapping language-image pre-training with frozen image encoders and large language models.
\newblock In \emph{ICML}, pages 19730--19742, 2023{\natexlab{b}}.

\bibitem[Li et~al.(2022{\natexlab{b}})Li, Ge, Yi, Hu, Shan, and Duan]{VLM:mc-beit}
Xiaotong Li, Yixiao Ge, Kun Yi, Zixuan Hu, Ying Shan, and Ling-Yu Duan.
\newblock mc-beit: Multi-choice discretization for image bert pre-training.
\newblock In \emph{ECCV}, pages 231--246, 2022{\natexlab{b}}.

\bibitem[Li et~al.(2024{\natexlab{c}})Li, Qiu, Chen, Kuen, Gu, Raj, and Lin]{li2024imagefolder}
Xiang Li, Kai Qiu, Hao Chen, Jason Kuen, Jiuxiang Gu, Bhiksha Raj, and Zhe Lin.
\newblock Imagefolder: Autoregressive image generation with folded tokens.
\newblock \emph{arXiv preprint arXiv:2410.01756}, 2024{\natexlab{c}}.

\bibitem[Li et~al.(2024{\natexlab{d}})Li, Zhang, Diao, Wang, Wang, and Duan]{VLM:Densefusion}
Xiaotong Li, Fan Zhang, Haiwen Diao, Yueze Wang, Xinlong Wang, and Ling-Yu Duan.
\newblock Densefusion-1m: Merging vision experts for comprehensive multimodal perception.
\newblock \emph{arXiv preprint arXiv:2407.08303}, 2024{\natexlab{d}}.

\bibitem[Li et~al.(2023{\natexlab{c}})Li, Du, Zhou, Wang, Zhao, and Wen]{Datasets:POPE}
Yifan Li, Yifan Du, Kun Zhou, Jinpeng Wang, Wayne~Xin Zhao, and Ji{-}Rong Wen.
\newblock Evaluating object hallucination in large vision-language models.
\newblock In \emph{EMNLP}, pages 292--305, 2023{\natexlab{c}}.

\bibitem[Li et~al.(2023{\natexlab{d}})Li, Yang, Liu, Ma, Zhang, Yang, Sun, Liu, and Bai]{VLM:Monkey}
Zhang Li, Biao Yang, Qiang Liu, Zhiyin Ma, Shuo Zhang, Jingxu Yang, Yabo Sun, Yuliang Liu, and Xiang Bai.
\newblock Monkey: Image resolution and text label are important things for large multi-modal models.
\newblock \emph{arXiv: 2311.06607}, 2023{\natexlab{d}}.

\bibitem[Lin et~al.(2024)Lin, Shrivastava, Luo, Iyer, Lewis, Gosh, Zettlemoyer, and Aghajanyan]{lin2024moma}
Xi~Victoria Lin, Akshat Shrivastava, Liang Luo, Srinivasan Iyer, Mike Lewis, Gargi Gosh, Luke Zettlemoyer, and Armen Aghajanyan.
\newblock Moma: Efficient early-fusion pre-training with mixture of modality-aware experts.
\newblock \emph{arXiv preprint arXiv:2407.21770}, 2024.

\bibitem[Liu et~al.(2023{\natexlab{a}})Liu, Li, Li, and Lee]{VLM:LLaVA-1.5}
Haotian Liu, Chunyuan Li, Yuheng Li, and Yong~Jae Lee.
\newblock Improved baselines with visual instruction tuning.
\newblock \emph{arXiv: 2310.03744}, 2023{\natexlab{a}}.

\bibitem[Liu et~al.(2023{\natexlab{b}})Liu, Li, Wu, and Lee]{VLM:LLaVA}
Haotian Liu, Chunyuan Li, Qingyang Wu, and Yong~Jae Lee.
\newblock Visual instruction tuning.
\newblock In \emph{NeurIPS}, 2023{\natexlab{b}}.

\bibitem[Liu et~al.(2024)Liu, Li, Li, Li, Zhang, Shen, and Lee]{VLM:LLaVA-1.6}
Haotian Liu, Chunyuan Li, Yuheng Li, Bo Li, Yuanhan Zhang, Sheng Shen, and Yong~Jae Lee.
\newblock Llava-next: Improved reasoning, ocr, and world knowledge, 2024.

\bibitem[Liu et~al.(2023{\natexlab{c}})Liu, Duan, Zhang, Li, Zhang, Zhao, Yuan, Wang, He, Liu, Chen, and Lin]{Datasets:MMBench}
Yuan Liu, Haodong Duan, Yuanhan Zhang, Bo Li, Songyang Zhang, Wangbo Zhao, Yike Yuan, Jiaqi Wang, Conghui He, Ziwei Liu, Kai Chen, and Dahua Lin.
\newblock Mmbench: Is your multi-modal model an all-around player?
\newblock \emph{arXiv: 2307.06281}, 2023{\natexlab{c}}.

\bibitem[Liu et~al.(2023{\natexlab{d}})Liu, Li, Yang, Li, Yin, Liu, Jin, and Bai]{Datasets:OCRBench}
Yuliang Liu, Zhang Li, Biao Yang, Chunyuan Li, Xucheng Yin, Cheng-lin Liu, Lianwen Jin, and Xiang Bai.
\newblock On the hidden mystery of ocr in large multimodal models.
\newblock \emph{arXiv preprint arXiv:2305.07895}, 2023{\natexlab{d}}.

\bibitem[Lu et~al.(2024)Lu, Liu, Zhang, Wang, Dong, Liu, Sun, Ren, Li, Yang, Sun, Deng, Xu, Xie, and Ruan]{VLM:DeepSeek-VL}
Haoyu Lu, Wen Liu, Bo Zhang, Bingxuan Wang, Kai Dong, Bo Liu, Jingxiang Sun, Tongzheng Ren, Zhuoshu Li, Hao Yang, Yaofeng Sun, Chengqi Deng, Hanwei Xu, Zhenda Xie, and Chong Ruan.
\newblock Deepseek-vl: Towards real-world vision-language understanding.
\newblock \emph{arXiv: 2403.05525}, 2024.

\bibitem[Lu et~al.(2022)Lu, Mishra, Xia, Qiu, Chang, Zhu, Tafjord, Clark, and Kalyan]{Datasets:ScienceQA}
Pan Lu, Swaroop Mishra, Tony Xia, Liang Qiu, Kai{-}Wei Chang, Song{-}Chun Zhu, Oyvind Tafjord, Peter Clark, and Ashwin Kalyan.
\newblock Learn to explain: Multimodal reasoning via thought chains for science question answering.
\newblock In \emph{NeurIPS}, 2022.

\bibitem[Luo et~al.(2024)Luo, Yang, Dou, Wang, Dai, Qiao, and Zhu]{luo2024mono}
Gen Luo, Xue Yang, Wenhan Dou, Zhaokai Wang, Jifeng Dai, Yu Qiao, and Xizhou Zhu.
\newblock Mono-internvl: Pushing the boundaries of monolithic multimodal large language models with endogenous visual pre-training.
\newblock \emph{arXiv preprint arXiv:2410.08202}, 2024.

\bibitem[Masry et~al.(2022)Masry, Do, Tan, Joty, and Hoque]{Datasets:ChartQA}
Ahmed Masry, Xuan~Long Do, Jia~Qing Tan, Shafiq Joty, and Enamul Hoque.
\newblock Chartqa: A benchmark for question answering about charts with visual and logical reasoning.
\newblock In \emph{ACL}, pages 2263--2279, 2022.

\bibitem[OpenAI(2023)]{VLM:GPT-4}
OpenAI.
\newblock {GPT-4} technical report.
\newblock \emph{arXiv: 2303.08774}, 2023.

\bibitem[Qu et~al.(2024)Qu, Zhang, Liu, Wang, Jiang, Gao, Ye, Du, Yuan, and Wu]{qu2024tokenflow}
Liao Qu, Huichao Zhang, Yiheng Liu, Xu Wang, Yi Jiang, Yiming Gao, Hu Ye, Daniel~K Du, Zehuan Yuan, and Xinglong Wu.
\newblock Tokenflow: Unified image tokenizer for multimodal understanding and generation.
\newblock \emph{arXiv preprint arXiv:2412.03069}, 2024.

\bibitem[Radford et~al.(2021)Radford, Kim, Hallacy, Ramesh, Goh, Agarwal, Sastry, Askell, Mishkin, Clark, Krueger, and Sutskever]{VLP:CLIP}
Alec Radford, Jong~Wook Kim, Chris Hallacy, Aditya Ramesh, Gabriel Goh, Sandhini Agarwal, Girish Sastry, Amanda Askell, Pamela Mishkin, Jack Clark, Gretchen Krueger, and Ilya Sutskever.
\newblock Learning transferable visual models from natural language supervision.
\newblock In \emph{ICML}, pages 8748--8763, 2021.

\bibitem[Schuhmann et~al.(2022)Schuhmann, Beaumont, Vencu, Gordon, Wightman, Cherti, Coombes, Katta, Mullis, Wortsman, et~al.]{Datasets:Laion-5b}
Christoph Schuhmann, Romain Beaumont, Richard Vencu, Cade Gordon, Ross Wightman, Mehdi Cherti, Theo Coombes, Aarush Katta, Clayton Mullis, Mitchell Wortsman, et~al.
\newblock Laion-5b: An open large-scale dataset for training next generation image-text models.
\newblock \emph{NeurIPS}, 35:\penalty0 25278--25294, 2022.

\bibitem[Singh et~al.(2019)Singh, Natarajan, Shah, Jiang, Chen, Batra, Parikh, and Rohrbach]{Datasets:TextVQA}
Amanpreet Singh, Vivek Natarajan, Meet Shah, Yu Jiang, Xinlei Chen, Dhruv Batra, Devi Parikh, and Marcus Rohrbach.
\newblock Towards {VQA} models that can read.
\newblock In \emph{CVPR}, 2019.

\bibitem[Sun et~al.(2024{\natexlab{a}})Sun, Jiang, Chen, Zhang, Peng, Luo, and Yuan]{LLamagen}
Peize Sun, Yi Jiang, Shoufa Chen, Shilong Zhang, Bingyue Peng, Ping Luo, and Zehuan Yuan.
\newblock Autoregressive model beats diffusion: Llama for scalable image generation.
\newblock \emph{arXiv preprint arXiv:2406.06525}, 2024{\natexlab{a}}.

\bibitem[Sun et~al.(2023{\natexlab{a}})Sun, Cui, Zhang, Zhang, Yu, Luo, Wang, Rao, Liu, Huang, and Wang]{VLM:EMUv2}
Quan Sun, Yufeng Cui, Xiaosong Zhang, Fan Zhang, Qiying Yu, Zhengxiong Luo, Yueze Wang, Yongming Rao, Jingjing Liu, Tiejun Huang, and Xinlong Wang.
\newblock Generative multimodal models are in-context learners.
\newblock \emph{arXiv: 2312.13286}, 2023{\natexlab{a}}.

\bibitem[Sun et~al.(2023{\natexlab{b}})Sun, Fang, Wu, Wang, and Cao]{TransF:EVA-CLIP}
Quan Sun, Yuxin Fang, Ledell Wu, Xinlong Wang, and Yue Cao.
\newblock {EVA-CLIP:} improved training techniques for {CLIP} at scale.
\newblock \emph{arXiv: 2303.15389}, 2023{\natexlab{b}}.

\bibitem[Sun et~al.(2023{\natexlab{c}})Sun, Yu, Cui, Zhang, Zhang, Wang, Gao, Liu, Huang, and Wang]{VLM:EMU}
Quan Sun, Qiying Yu, Yufeng Cui, Fan Zhang, Xiaosong Zhang, Yueze Wang, Hongcheng Gao, Jingjing Liu, Tiejun Huang, and Xinlong Wang.
\newblock Generative pretraining in multimodality.
\newblock \emph{arXiv: 2307.05222}, 2023{\natexlab{c}}.

\bibitem[Sun et~al.(2024{\natexlab{b}})Sun, Wang, Yu, Cui, Zhang, Zhang, and Wang]{TransF:EVA-CLIP-18B}
Quan Sun, Jinsheng Wang, Qiying Yu, Yufeng Cui, Fan Zhang, Xiaosong Zhang, and Xinlong Wang.
\newblock {EVA-CLIP-18B:} scaling {CLIP} to 18 billion parameters.
\newblock \emph{arXiv: 2402.04252}, 2024{\natexlab{b}}.

\bibitem[Team(2024{\natexlab{a}})]{team2024chameleon}
Chameleon Team.
\newblock Chameleon: Mixed-modal early-fusion foundation models.
\newblock \emph{arXiv preprint arXiv:2405.09818}, 2024{\natexlab{a}}.

\bibitem[Team et~al.(2023)Team, Anil, Borgeaud, Wu, Alayrac, Yu, Soricut, Schalkwyk, Dai, Hauth, et~al.]{VLM:Gemini}
Gemini Team, Rohan Anil, Sebastian Borgeaud, Yonghui Wu, Jean-Baptiste Alayrac, Jiahui Yu, Radu Soricut, Johan Schalkwyk, Andrew~M Dai, Anja Hauth, et~al.
\newblock Gemini: a family of highly capable multimodal models.
\newblock \emph{arXiv: 2312.11805}, 2023.

\bibitem[Team(2023)]{TransF:InternLM}
InternLM Team.
\newblock Internlm: A multilingual language model with progressively enhanced capabilities.
\newblock \url{https://github.com/InternLM/InternLM}, 2023.

\bibitem[Team(2024{\natexlab{b}})]{VLMs:LLama3.2}
Meta Team.
\newblock Llama 3.2: Revolutionizing edge ai and vision with open, customizable models, 2024{\natexlab{b}}.

\bibitem[Team(2024{\natexlab{c}})]{qwen2.5}
Qwen Team.
\newblock Qwen2.5: A party of foundation models, 2024{\natexlab{c}}.

\bibitem[Tong et~al.(2024{\natexlab{a}})Tong, Brown, Wu, Woo, Middepogu, Akula, Yang, Yang, Iyer, Pan, et~al.]{tong2024cambrian}
Shengbang Tong, Ellis Brown, Penghao Wu, Sanghyun Woo, Manoj Middepogu, Sai~Charitha Akula, Jihan Yang, Shusheng Yang, Adithya Iyer, Xichen Pan, et~al.
\newblock Cambrian-1: A fully open, vision-centric exploration of multimodal llms.
\newblock \emph{arXiv preprint arXiv:2406.16860}, 2024{\natexlab{a}}.

\bibitem[Tong et~al.(2024{\natexlab{b}})Tong, Liu, Zhai, Ma, LeCun, and Xie]{tong2024eyes}
Shengbang Tong, Zhuang Liu, Yuexiang Zhai, Yi Ma, Yann LeCun, and Saining Xie.
\newblock Eyes wide shut? exploring the visual shortcomings of multimodal llms.
\newblock In \emph{CVPR}, pages 9568--9578, 2024{\natexlab{b}}.

\bibitem[Touvron et~al.(2023{\natexlab{a}})Touvron, Lavril, Izacard, Martinet, Lachaux, Lacroix, Rozi{\`{e}}re, Goyal, Hambro, Azhar, Rodriguez, Joulin, Grave, and Lample]{TransF:LLaMA}
Hugo Touvron, Thibaut Lavril, Gautier Izacard, Xavier Martinet, Marie{-}Anne Lachaux, Timoth{\'{e}}e Lacroix, Baptiste Rozi{\`{e}}re, Naman Goyal, Eric Hambro, Faisal Azhar, Aur{\'{e}}lien Rodriguez, Armand Joulin, Edouard Grave, and Guillaume Lample.
\newblock Llama: Open and efficient foundation language models.
\newblock \emph{arXiv: 2302.13971}, 2023{\natexlab{a}}.

\bibitem[Touvron et~al.(2023{\natexlab{b}})Touvron, Martin, Stone, Albert, Almahairi, Babaei, Bashlykov, Batra, Bhargava, Bhosale, et~al.]{TransF:LLaMA2}
Hugo Touvron, Louis Martin, Kevin Stone, Peter Albert, Amjad Almahairi, Yasmine Babaei, Nikolay Bashlykov, Soumya Batra, Prajjwal Bhargava, Shruti Bhosale, et~al.
\newblock Llama 2: Open foundation and fine-tuned chat models.
\newblock \emph{arXiv: 2307.09288}, 2023{\natexlab{b}}.

\bibitem[Van Den~Oord et~al.(2017)Van Den~Oord, Vinyals, et~al.]{van2017VQVAE}
Aaron Van Den~Oord, Oriol Vinyals, et~al.
\newblock Neural discrete representation learning.
\newblock \emph{NeurIPS}, 30, 2017.

\bibitem[Vaswani et~al.(2017)Vaswani, Shazeer, Parmar, Uszkoreit, Jones, Gomez, Kaiser, and Polosukhin]{TransF:Transformer}
Ashish Vaswani, Noam Shazeer, Niki Parmar, Jakob Uszkoreit, Llion Jones, Aidan~N. Gomez, Lukasz Kaiser, and Illia Polosukhin.
\newblock Attention is all you need.
\newblock In \emph{NIPS}, pages 5998--6008, 2017.

\bibitem[Wang et~al.(2023)Wang, Meng, Weng, He, Wu, and Jiang]{VLM:LVIS-4v}
Junke Wang, Lingchen Meng, Zejia Weng, Bo He, Zuxuan Wu, and Yu{-}Gang Jiang.
\newblock To see is to believe: Prompting {GPT-4V} for better visual instruction tuning.
\newblock \emph{arXiv: 2311.07574}, 2023.

\bibitem[Wang et~al.(2024{\natexlab{a}})Wang, Bai, Tan, Wang, Fan, Bai, Chen, Liu, Wang, Ge, et~al.]{wang2024qwen2-vl}
Peng Wang, Shuai Bai, Sinan Tan, Shijie Wang, Zhihao Fan, Jinze Bai, Keqin Chen, Xuejing Liu, Jialin Wang, Wenbin Ge, et~al.
\newblock Qwen2-vl: Enhancing vision-language model's perception of the world at any resolution.
\newblock \emph{arXiv preprint arXiv:2409.12191}, 2024{\natexlab{a}}.

\bibitem[Wang et~al.(2022)Wang, Bao, Dong, Bjorck, Peng, Liu, Aggarwal, Mohammed, Singhal, Som, and Wei]{VLP:BEiTv3}
Wenhui Wang, Hangbo Bao, Li Dong, Johan Bjorck, Zhiliang Peng, Qiang Liu, Kriti Aggarwal, Owais~Khan Mohammed, Saksham Singhal, Subhojit Som, and Furu Wei.
\newblock Image as a foreign language: Beit pretraining for all vision and vision-language tasks.
\newblock \emph{arXiv: 2208.10442}, 2022.

\bibitem[Wang et~al.(2024{\natexlab{b}})Wang, Zhang, Luo, Sun, Cui, Wang, Zhang, Wang, Li, Yu, et~al.]{VLM:EMU3}
Xinlong Wang, Xiaosong Zhang, Zhengxiong Luo, Quan Sun, Yufeng Cui, Jinsheng Wang, Fan Zhang, Yueze Wang, Zhen Li, Qiying Yu, et~al.
\newblock Emu3: Next-token prediction is all you need.
\newblock \emph{arXiv preprint arXiv:2409.18869}, 2024{\natexlab{b}}.

\bibitem[Wang et~al.(2024{\natexlab{c}})Wang, Zhu, Xu, Zhou, Liu, Zhang, Wang, Shi, Li, Li, et~al.]{wang2024mio}
Zekun Wang, King Zhu, Chunpu Xu, Wangchunshu Zhou, Jiaheng Liu, Yibo Zhang, Jiashuo Wang, Ning Shi, Siyu Li, Yizhi Li, et~al.
\newblock Mio: A foundation model on multimodal tokens.
\newblock \emph{arXiv preprint arXiv:2409.17692}, 2024{\natexlab{c}}.

\bibitem[Wu et~al.(2024{\natexlab{a}})Wu, Chen, Wu, Ma, Liu, Pan, Liu, Xie, Yu, Ruan, et~al.]{wu2024janus}
Chengyue Wu, Xiaokang Chen, Zhiyu Wu, Yiyang Ma, Xingchao Liu, Zizheng Pan, Wen Liu, Zhenda Xie, Xingkai Yu, Chong Ruan, et~al.
\newblock Janus: Decoupling visual encoding for unified multimodal understanding and generation.
\newblock \emph{arXiv preprint arXiv:2410.13848}, 2024{\natexlab{a}}.

\bibitem[Wu et~al.(2024{\natexlab{b}})Wu, Zhang, Chen, Tang, Li, Fang, Zhu, Xie, Yin, Yi, et~al.]{wu2024vila-u}
Yecheng Wu, Zhuoyang Zhang, Junyu Chen, Haotian Tang, Dacheng Li, Yunhao Fang, Ligeng Zhu, Enze Xie, Hongxu Yin, Li Yi, et~al.
\newblock Vila-u: a unified foundation model integrating visual understanding and generation.
\newblock \emph{arXiv preprint arXiv:2409.04429}, 2024{\natexlab{b}}.

\bibitem[x.ai(2024)]{Datasets:Realworldqa}
x.ai.
\newblock Grok-1.5 vision preview, 2024.

\bibitem[Xie et~al.(2024{\natexlab{a}})Xie, Mao, Bai, Zhang, Wang, Lin, Gu, Chen, Yang, and Shou]{xie2024show-o}
Jinheng Xie, Weijia Mao, Zechen Bai, David~Junhao Zhang, Weihao Wang, Kevin~Qinghong Lin, Yuchao Gu, Zhijie Chen, Zhenheng Yang, and Mike~Zheng Shou.
\newblock Show-o: One single transformer to unify multimodal understanding and generation.
\newblock \emph{arXiv preprint arXiv:2408.12528}, 2024{\natexlab{a}}.

\bibitem[Xie et~al.(2024{\natexlab{b}})Xie, Du, Song, and Liu]{xie2024muse}
Rongchang Xie, Chen Du, Ping Song, and Chang Liu.
\newblock Muse-vl: Modeling unified vlm through semantic discrete encoding.
\newblock \emph{arXiv preprint arXiv:2411.17762}, 2024{\natexlab{b}}.

\bibitem[Xu et~al.(2024)Xu, Yao, Guo, Cui, Ni, Ge, Chua, Liu, Sun, and Huang]{VLM:LLaVA-UHD}
Ruyi Xu, Yuan Yao, Zonghao Guo, Junbo Cui, Zanlin Ni, Chunjiang Ge, Tat{-}Seng Chua, Zhiyuan Liu, Maosong Sun, and Gao Huang.
\newblock Llava-uhd: an {LMM} perceiving any aspect ratio and high-resolution images.
\newblock \emph{arXiv: 2403.11703}, 2024.

\bibitem[Xue et~al.(2024)Xue, Shu, Awadalla, Wang, Yan, Purushwalkam, Zhou, Prabhu, Dai, Ryoo, et~al.]{xue2024blip-3}
Le Xue, Manli Shu, Anas Awadalla, Jun Wang, An Yan, Senthil Purushwalkam, Honglu Zhou, Viraj Prabhu, Yutong Dai, Michael~S Ryoo, et~al.
\newblock xgen-mm (blip-3): A family of open large multimodal models.
\newblock \emph{arXiv preprint arXiv:2408.08872}, 2024.

\bibitem[Yang et~al.(2024)Yang, Yang, Hui, Zheng, Yu, Zhou, Li, Li, Liu, Huang, et~al.]{yang2024qwen2}
An Yang, Baosong Yang, Binyuan Hui, Bo Zheng, Bowen Yu, Chang Zhou, Chengpeng Li, Chengyuan Li, Dayiheng Liu, Fei Huang, et~al.
\newblock Qwen2 technical report.
\newblock \emph{arXiv preprint arXiv:2407.10671}, 2024.

\bibitem[Yang et~al.(2023)Yang, Li, Lin, Wang, Lin, Liu, and Wang]{VLM:GPT-4v}
Zhengyuan Yang, Linjie Li, Kevin Lin, Jianfeng Wang, Chung-Ching Lin, Zicheng Liu, and Lijuan Wang.
\newblock The dawn of lmms: Preliminary explorations with gpt-4v (ision).
\newblock \emph{arXiv: 2309.17421}, 9, 2023.

\bibitem[Ye et~al.(2023{\natexlab{a}})Ye, Xu, Xu, Ye, Yan, Zhou, Wang, Hu, Shi, Shi, Li, Xu, Chen, Tian, Qi, Zhang, and Huang]{VLM:mPLUG-Owl}
Qinghao Ye, Haiyang Xu, Guohai Xu, Jiabo Ye, Ming Yan, Yiyang Zhou, Junyang Wang, Anwen Hu, Pengcheng Shi, Yaya Shi, Chenliang Li, Yuanhong Xu, Hehong Chen, Junfeng Tian, Qian Qi, Ji Zhang, and Fei Huang.
\newblock mplug-owl: Modularization empowers large language models with multimodality.
\newblock \emph{arXiv: 2304.14178}, 2023{\natexlab{a}}.

\bibitem[Ye et~al.(2023{\natexlab{b}})Ye, Xu, Ye, Yan, Hu, Liu, Qian, Zhang, Huang, and Zhou]{VLM:mPLUG-Owl2}
Qinghao Ye, Haiyang Xu, Jiabo Ye, Ming Yan, Anwen Hu, Haowei Liu, Qi Qian, Ji Zhang, Fei Huang, and Jingren Zhou.
\newblock mplug-owl2: Revolutionizing multi-modal large language model with modality collaboration.
\newblock \emph{arXiv: 2311.04257}, 2023{\natexlab{b}}.

\bibitem[Yu et~al.(2023)Yu, Yang, Li, Wang, Lin, Liu, Wang, and Wang]{Datasets:MM-vet}
Weihao Yu, Zhengyuan Yang, Linjie Li, Jianfeng Wang, Kevin Lin, Zicheng Liu, Xinchao Wang, and Lijuan Wang.
\newblock Mm-vet: Evaluating large multimodal models for integrated capabilities.
\newblock \emph{arXiv: 2308.02490}, 2023.

\bibitem[Yue et~al.(2023)Yue, Ni, Zhang, Zheng, Liu, Zhang, Stevens, Jiang, Ren, Sun, et~al.]{Datasets:MMMU}
Xiang Yue, Yuansheng Ni, Kai Zhang, Tianyu Zheng, Ruoqi Liu, Ge Zhang, Samuel Stevens, Dongfu Jiang, Weiming Ren, Yuxuan Sun, et~al.
\newblock Mmmu: A massive multi-discipline multimodal understanding and reasoning benchmark for expert agi.
\newblock \emph{arXiv: 2311.16502}, 2023.

\bibitem[Zhai et~al.(2023)Zhai, Mustafa, Kolesnikov, and Beyer]{TransF:Siglip}
Xiaohua Zhai, Basil Mustafa, Alexander Kolesnikov, and Lucas Beyer.
\newblock Sigmoid loss for language image pre-training.
\newblock In \emph{ICCV}, pages 11975--11986, 2023.

\bibitem[Zhan et~al.(2024)Zhan, Dai, Ye, Zhou, Zhang, Liu, Zhang, Yuan, Zhang, Li, Yan, Fu, Gui, Sun, Jiang, and Qiu]{VLM:AnyGPT}
Jun Zhan, Junqi Dai, Jiasheng Ye, Yunhua Zhou, Dong Zhang, Zhigeng Liu, Xin Zhang, Ruibin Yuan, Ge Zhang, Linyang Li, Hang Yan, Jie Fu, Tao Gui, Tianxiang Sun, Yugang Jiang, and Xipeng Qiu.
\newblock Anygpt: Unified multimodal {LLM} with discrete sequence modeling.
\newblock \emph{arXiv: 2402.12226}, 2024.

\bibitem[Zhang et~al.(2024{\natexlab{a}})Zhang, Li, Zhang, Pu, Cahyono, Hu, Liu, Zhang, Yang, Li, et~al.]{zhang2024lmms-eval}
Kaichen Zhang, Bo Li, Peiyuan Zhang, Fanyi Pu, Joshua~Adrian Cahyono, Kairui Hu, Shuai Liu, Yuanhan Zhang, Jingkang Yang, Chunyuan Li, et~al.
\newblock Lmms-eval: Reality check on the evaluation of large multimodal models.
\newblock \emph{arXiv preprint arXiv:2407.12772}, 2024{\natexlab{a}}.

\bibitem[Zhang et~al.(2024{\natexlab{b}})Zhang, Han, Zhou, Hu, Yan, Lu, Li, Gao, and Qiao]{VLM:Llama-Adapter}
Renrui Zhang, Jiaming Han, Aojun Zhou, Xiangfei Hu, Shilin Yan, Pan Lu, Hongsheng Li, Peng Gao, and Yu Qiao.
\newblock Llama-adapter: Efficient fine-tuning of language models with zero-init attention.
\newblock In \emph{ICLR}, 2024{\natexlab{b}}.

\bibitem[Zheng et~al.(2022)Zheng, Vuong, Cai, and Phung]{VLM:movq}
Chuanxia Zheng, Tung-Long Vuong, Jianfei Cai, and Dinh Phung.
\newblock Movq: Modulating quantized vectors for high-fidelity image generation.
\newblock \emph{NeurIPS}, 35:\penalty0 23412--23425, 2022.

\bibitem[Zheng et~al.(2023)Zheng, Yin, Xie, Huang, Sun, Yu, Cao, Kozyrakis, Stoica, Gonzalez, Barrett, and Sheng]{VLM:SGlang}
Lianmin Zheng, Liangsheng Yin, Zhiqiang Xie, Jeff Huang, Chuyue Sun, Cody~Hao Yu, Shiyi Cao, Christos Kozyrakis, Ion Stoica, Joseph~E. Gonzalez, Clark~W. Barrett, and Ying Sheng.
\newblock Efficiently programming large language models using sglang.
\newblock \emph{arXiv: 2312.07104}, 2023.

\end{thebibliography}
}

\clearpage
\begin{appendix}

\twocolumn[{
\renewcommand\twocolumn[1][]{#1}
\maketitle 
\vspace{-8mm}
\begin{center} 
\centering 
\captionof{table}{Experiment Details in the main body. Note that T.M. denotes trainable modules in each stage. PEL, PAL, and VLayers represent patch embedding, alignment, and newly added vision layers within the LLM. EVE-recap-8/29M indicates a subset 8M of 29M training data.}
\resizebox{\linewidth}{!}{
\begin{tabular}{l|cc|cc|cccc|cc}
\toprule
\multirow{2}{*}{\textbf{Exp.}} 
&\multirow{2}{*}{\textbf{Model}}
&\multirow{2}{*}{\textbf{LLM}}
&\multicolumn{2}{c|}{\textbf{Stage 1}} 
&\multicolumn{4}{c|}{\textbf{Stage 2}}
&\multicolumn{2}{c}{\textbf{Stage 3}}\\
& &
&\textbf{Data} &\textbf{T.M.}
&\multicolumn{2}{c}{\textbf{Training Data}} 
&\multicolumn{2}{c|}{\textbf{Trainable Module}}
&\textbf{Data} &\textbf{T.M.} \\
\midrule
Fig.2 (i)
& EVEv1.0 & Vicuna-7B
& EVE-cap-16M & PEL
& \multicolumn{2}{c}{EVE-cap-33M} 
& \multicolumn{2}{c|}{PEL, LLM}
& LLaVA-mix-665k & PEL, LLM
\\
\hdashline
\multirow{4}{*}{Fig.5}
&EVEv1.0 & Qwen2.5-7B
& EVE-recap-10M & PEL
& \multicolumn{2}{c}{EVE-recap-8/29M} 
& \multicolumn{2}{c|}{PEL}
& LLaVA-mix-665k & PEL, LLM
\\
&EVEv1.2 & Qwen2.5-7B
& EVE-recap-10M & PEL
& \multicolumn{2}{c}{EVE-recap-8/29M} 
& \multicolumn{2}{c|}{PEL, VLayers}
& LLaVA-mix-665k & PEL, LLM\\
&EVEv1.5 & Qwen2.5-7B
& EVE-recap-10M & PEL
& \multicolumn{2}{c}{EVE-recap-8/29M} 
& \multicolumn{2}{c|}{PEL, VLayers}
& LLaVA-mix-665k & PEL, LLM\\
&EVEv2.0 & Qwen2.5-7B
& EVE-recap-10M & PEL
& \multicolumn{2}{c}{EVE-recap-8/29M} 
& \multicolumn{2}{c|}{PEL, VLayers}
& LLaVA-mix-665k & PEL, LLM
\\
\hdashline
Fig.6 
& EVEv1.0 & Vicuna-7B
& 10M varied data & PEL
& \multicolumn{2}{c}{8M same data from Stage 1}
& \multicolumn{2}{c|}{PEL, LLM}
& LLaVA-mix-665k & PEL, LLM
\\
\hdashline
\multirow{4}{*}{Fig.7}
&EVE [17] &Qwen2.5-7B
&EVE-cap/recap-10/48M &PEL, PAL
&\multicolumn{2}{c}{EVE-cap/recap-48M} 
&\multicolumn{2}{c|}{PEL, PAL, LLM}
&EVE-sft-7M &PEL, PAL, LLM
\\
&Mono-InternVL [50] &Qwen2.5-7B
&EVE-cap/recap-10/48M &PEL
&\multicolumn{2}{c}{EVE-cap/recap-48M} 
&\multicolumn{2}{c|}{PEL, VLayers}
&EVE-sft-7M &PEL, LLM
\\
&EVEv2.0 &Qwen2.5-7B
&EVE-cap/recap-10/48M &PEL
&\multicolumn{2}{c}{EVE-cap/recap-48M} 
&\multicolumn{2}{c|}{PEL, VLayers}
&EVE-sft-7M &PEL, LLM
\\
&LLaVA-1.5 [43] &Qwen2.5-7B
&EVE-cap/recap-10/48M &Projector
&\multicolumn{2}{c}{EVE-cap/recap-48M} 
&\multicolumn{2}{c|}{Vision Encoder, Projector}
&EVE-sft-7M &All Layers
\\
\hdashline
Fig.8
& EVEv1.0 & Vicuna-7B
& 10M varied data & PEL
& \multicolumn{2}{c}{8M same data from Stage 1}
& \multicolumn{2}{c|}{PEL, LLM}
& LLaVA-mix-665k & PEL, LLM
\\
\hdashline
Supp.Fig.1
& EVEv1.0 & Vicuna-7B
& EVE-recap-10M & PEL
& \multicolumn{2}{c}{EVE-recap-8/29M}
& \multicolumn{2}{c|}{PEL, LLM}
& LLaVA-mix-665k & PEL, LLM
\\
\bottomrule
\multirow{2}{*}{\textbf{Exp.}} 
&\multirow{2}{*}{\textbf{Model}}
&\multirow{2}{*}{\textbf{LLM}}
&\multicolumn{2}{c|}{\textbf{Stage 1}} 
&\multicolumn{2}{c}{\textbf{Stage 2.1}}
&\multicolumn{2}{c|}{\textbf{Stage 2.2}}
&\multicolumn{2}{c}{\textbf{Stage 3}}\\
& &
&\textbf{Data} &\textbf{T.M.}
&\textbf{Data} &\textbf{T.M.}
&\textbf{Data} &\textbf{T.M.}
&\textbf{Data} &\textbf{T.M.} \\
\midrule
\multirow{2}{*}{Fig.2 (ii)}
& EVEv1.0 & Qwen2-7B
& EVE-recap-10M & PEL
& EVE-recap-29M & PEL
& EVE-recap-48M & PEL, LLM
& Various SFT data & PEL, LLM
\\
& EVEv1.2 & Qwen2-7B
& EVE-recap-10M & PEL
& EVE-recap-29M & PEL, VLayers
& EVE-recap-48M & PEL, LLM
& Various SFT data & PEL, LLM 
\\
\hdashline
Tab.2 
& EVEv2.0 &Qwen2.5-7B
& EVE-recap-10M & PEL
& EVE-recap-77M & PEL, VLayers
& EVE-multi-task-5M & PEL, LLM
& EVE-sft-7M & PEL, LLM  
\\
\hdashline
Fig.9
& EVEv2.0 &Qwen2.5-7B
& EVE-recap-10M & PEL
& EVE-recap-77M & PEL, VLayers
& EVE-multi-task-5M & PEL, LLM
& EVE-sft-7M & PEL, LLM  
\\
\bottomrule
\end{tabular}}
\label{tab:exp_setting}
\end{center}
}]

\section{Experiment Details}
\label{sec:exp-details}

All experiment details in the main body are listed in~\Cref{tab:exp_setting}. ``EVE-cap-16M'' denotes the data mixture of LAION~\cite{Datasets:Laion-5b}, OpenImages~\cite{Datasets:OpenImages}, and SAM~\cite{TransF:SAM} annotated by LLaVA-1.5 (13B) and Emu2 (17B). ``EVE-recap-16M'' denotes the data mixture of Datacomp~\cite{Datasets:datacomp}, LAION~\cite{Datasets:Laion-5b}, OpenImages~\cite{Datasets:OpenImages}, and SAM~\cite{TransF:SAM} annotated by DenseFusion++ (7B).

For Exp.(i), we first use EVE-cap-16M in Stage 1 to train the projector, vision vocabulary embeddings, and lightweight vision block for the vision encoder (\textit{VE}), the discrete tokenizer (\textit{DT}), and EVEv1.0. In Stage 2, we use EVE-cap-33M to train only the vision encoder and projector for \textit{VE}, as unfreezing LLM weights at this stage leads to performance collapse~\cite{VLM:EVE}. For \textit{DT} and EVEv1.0, we unfreeze all model parameters.
In Stage 3, we train all model weights across all models. Finally, we quantify weight changes between Vicuna-7B~\cite{TransF:Vicuna} and \textit{VE} / EVEv1.0 to analyze the architectural differences. To ensure fairness, we remove the original vision encoder supervision in EVEv1.0.

For Exp.(ii), all VLMs use stronger Qwen2-7B~\cite{yang2024qwen2} and high-quality EVE-recap. In Stage 1, we train the projector for \textit{VE}, patch embedding layer for EVEv1.0, patch embedding and extra vision layer inside the LLM for EVEv1.2. In Stage 2 and 3, we train all model weights for EVEv1.0-1.2. Note that we skip Stage 2 for \textit{VE} as the baseline for comparison. Here, we compare the weights between Qwen2 and \textit{VE} / EVEv1.2 trained by LLaVA-onevision~\cite{Llava-onevision}.

\begin{table}[t!]
\caption{Dataset details in Stage 2.2, and 3 for fine-tuning EVEv2.0. Note that ***-FL denotes the filtered training dataset.}
\centering
\resizebox{\linewidth}{!}{
\begin{tabular}{l|l|c}
\toprule
\textbf{Stage} & \textbf{Dataset} & \textbf{\#Data} \\
\midrule
\multirow{5}{*}{2.2} 
&Cambrian-FL~\cite{tong2024cambrian},
Infinity-Instruct-FL~\cite{gu2024infinity},
&  \multirow{5}{*}{5M}\\
& LVIS-Instruct-FL~\cite{VLM:LVIS-4v},
Sharegpt4v-FL~\cite{VLM:Sharegpt4v},
& \\
&ALLaVA-laion-FL~\cite{chen2024allava},
ALLaVA-vflan-FL~\cite{chen2024allava},
&  \\
&LLaVA-Pretrain-FL~\cite{VLM:LLaVA-1.5},
DocReason-FL~\cite{hu2024mplug-docowl2},
&  \\
&DocDownstream-FL~\cite{hu2024mplug-docowl2},
DocStruct4M-FL~\cite{hu2024mplug-docowl2}.
&  \\
\midrule
\multirow{4}{*}{3} 
& LLaVA-onevision~\cite{Llava-onevision}, 
Infinity-MM-Synthesis~\cite{gu2024infinity},
& \multirow{4}{*}{7.3M}  \\
&Infinity-MM-Preference~\cite{gu2024infinity},
Infinity-Instruct-FL~\cite{gu2024infinity},
& \\
&DenseFusion~\cite{VLM:Densefusion},
Cambrian-FL~\cite{tong2024cambrian},
Docmatix-FL~\cite{laurenccon2024docmatix},
& \\
&
LVIS-Instruct-FL~\cite{VLM:LVIS-4v},
BLIP-OCR\cite{VLM:InstructBLIP},
LLaVA-mix~\cite{VLM:LLaVA-1.5}.
&  \\
\bottomrule
\end{tabular}}
\label{tab:datasets_details}
\end{table}

\section{Dataset Details}
From~\Cref{tab:datasets_details}, we use Cambrian-FL~\cite{tong2024cambrian}, Infinity-Instruct-FL~\cite{gu2024infinity},
LVIS-Instruct-FL~\cite{VLM:LVIS-4v},
Sharegpt4v-FL~\cite{VLM:Sharegpt4v},
ALLaVA-laion-FL~\cite{chen2024allava},
ALLaVA-vflan-FL~\cite{chen2024allava},
LLaVA-Pretrain-FL~\cite{VLM:LLaVA-1.5}, 
DocReason-FL~\cite{hu2024mplug-docowl2},
DocDownstream-FL~\cite{hu2024mplug-docowl2},
and DocStruct4M-FL~\cite{hu2024mplug-docowl2} in Stage 2.2.
In Stage 3, we adopt LLaVA-onevision~\cite{Llava-onevision} and Infinity-MM~\cite{gu2024infinity}.

\section{Hyper-parameter Configurations}

The detailed implementation configurations in Stages 1, 2.1, 2.2, and 3 are summarized in~\Cref{tab:hyperparam}. 

\begin{table}[!t]
    \centering
    \caption{Hyper-parameter configurations in Stage 1-3 for training EVEv2.0. Note that we set the training epoch in each stage as 1.}
    \resizebox{\linewidth}{!}{
    \begin{tabular}{lcccc}
         \toprule
         Configuration & Stage 1 & Stage 2.1 & Stage 2.2 & Stage 3 \\
         \midrule
         Maximum Patch Token   & $625$ & $625-2500$ & $2500$ & $2500$ \\
         Optimizer & \multicolumn{4}{c}{AdamW} \\
         Hyperparameters & \multicolumn{4}{c}{$\beta_{1}=0.9, \beta_{2}=0.999, eps=1e^{-8}$} \\
         Peak learning rate       & $2e^{-4}$ & $1e^{-4}$ & $2e^{-5}$ & $1e^{-5}$\\
         LR schedule   & \multicolumn{4}{c}{cosine decay with warm-up}\\
         Warm-up steps  & \multicolumn{4}{c}{$0.03$}\\
         Weight decay             & \multicolumn{4}{c}{$0.0$} \\
         Global batch size        & $1024$ & $1024$ & $512$ & $512$\\
         Numerical precision      & \multicolumn{4}{c}{$\mathtt{bfloat16}$} \\
         \bottomrule
    \end{tabular}
    }
    \label{tab:hyperparam}
\end{table}

\section{Additional Ablation Studies}

\textbf{Decreasing SQA-Img metric with more training data.
} 
We empirically discovered that the performance of the SQA-Img metric is strongly influenced by the knowledge of the pre-trained LLM, making it prone to catastrophic linguistic forgetting issues. As it primarily involves text-centric knowledge tasks, increasing image captioning data alone degrades performance when the LLM parameters are unfrozen.

\textbf{EVEv1.2: lowest loss, poor performance in~\Cref{fig:different_version}.} 
This is because EVEv1.2 progressively updates the LLM weights while quickly adapting to the simplified language structures within the VL caption data.
Notably, the SFT losses—which better reflect task performance—exhibit a clear downward trend (EVEv1.2: 1.0306, EVEv1.5: 0.9513, EVEv2: 0.9327), indicating that EVEv1.2 is less effective at alleviating LLM forgetting and multimodal interference.

\textbf{Consistent benefits across different LLM capacities.} 
We train EVEv1, v1.5, and v2 using Qwen2.5-1.5B with EVE-recap-4M and LLaVA-mix-665k, achieving average scores of 35.1\%, 37.3\%, and 38.5\% across six diverse vision-language benchmarks. This further confirms that sparse design consistently benefits models of different scales.

\begin{figure}[!h]
    \centering 
    \includegraphics[width=\linewidth,trim= 0 0 0 20,clip]
    {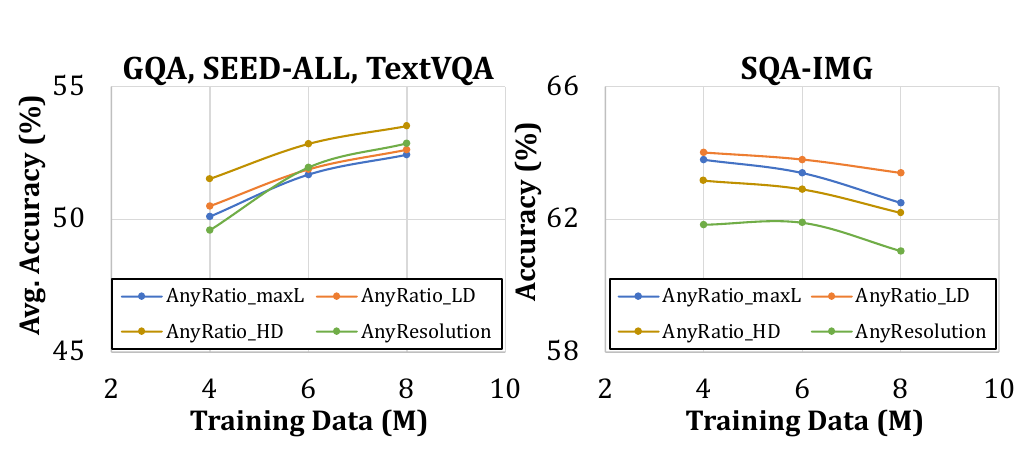} 
    \caption{Evaluation results of image settings. 
    We use EVEv1.0 with Vicuna-7B~\cite{TransF:Vicuna}.
    ``AnyRatio\_maxL'': longest image edge as 800, 
    ``AnyRatio\_LD'': fixed image area as 800$^2$,
    ``AnyRatio\_HD'': fixed image area as 1600$^2$,
    ``AnyResolution'': arbitrary resolution. }
    \label{fig:image_encoding}
\end{figure}

\textbf{Flexible image processing mode of encoder-free VLMs.}
In~\Cref{fig:image_encoding}, we explore four different input formats. Among them, ``AnyRatio\_HD'' (standard) provides the best performance gains, while ``AnyResolution'' performs poorly in the early stages but shows improved data-scaling efficiency over data scales. This early underperformance is likely due to limited pre-training data and the imbalance in image resolutions. We believe that with sufficient and well-balanced data, arbitrary-resolution inputs can offer better computational efficiency and greater flexibility for real-world images.

\begin{table}[h!]
    \centering
    \caption{Inference speed comparison of various EVE counterparts}
    \label{tab:TTFT_TPS}
    \resizebox{\linewidth}{!}{
    \begin{tabular}{l|cccc}
         \hline
         Model &EVEv1.0 & EVEv1.2 & EVEv1.5 & EVEv2.0 \\
         TTFT (s) &3.77 &3.77 &4.01 &4.25 \\
         \hline
    \end{tabular}}
\end{table}

\textbf{Inference speed comparisons across EVE variants.}
EVEv2.0 decouples weights across modalities at each layer, totaling 2$\times$7B parameters with active FLOPs per token matching a 7B dense model. 
Notably, we adopt an NVIDIA A100 with PyTorch, fixing image, input text, and output text tokens to 2500, 200, and 100, respectively. We report TTFT (time to first token) and TPS (tokens per second after the first token). 
Here, EVEv1.5 / EVEv2 bring marginal 6 / 13\% TTFT overheads, while all EVEv1–v2 models sustain 35 TPS (tok/s) using KV-cache for visual and text prefixes.

\section{Visual Understanding Demonstration}
We investigate several vision perception and reasoning capabilities of EVEv2.0, including OCR capability in~\Cref{tab:ocr_in_web,tab:ocr_in_ppt}, mixed information in~\Cref{tab:poster}, real-world scenes in~\Cref{tab:real1,tab:real2}, and set-of-mark prompting task in~\Cref{tab:refer_to_marks}.

\begin{table*}
\begin{minipage}{0.99\textwidth}
\begin{AIbox}{Example 1: Understanding OCR Information in Webpage}
\centering
\scalebox{0.80}{
\begin{tabular}{l p{18cm}}
&  \includegraphics[width=16cm]{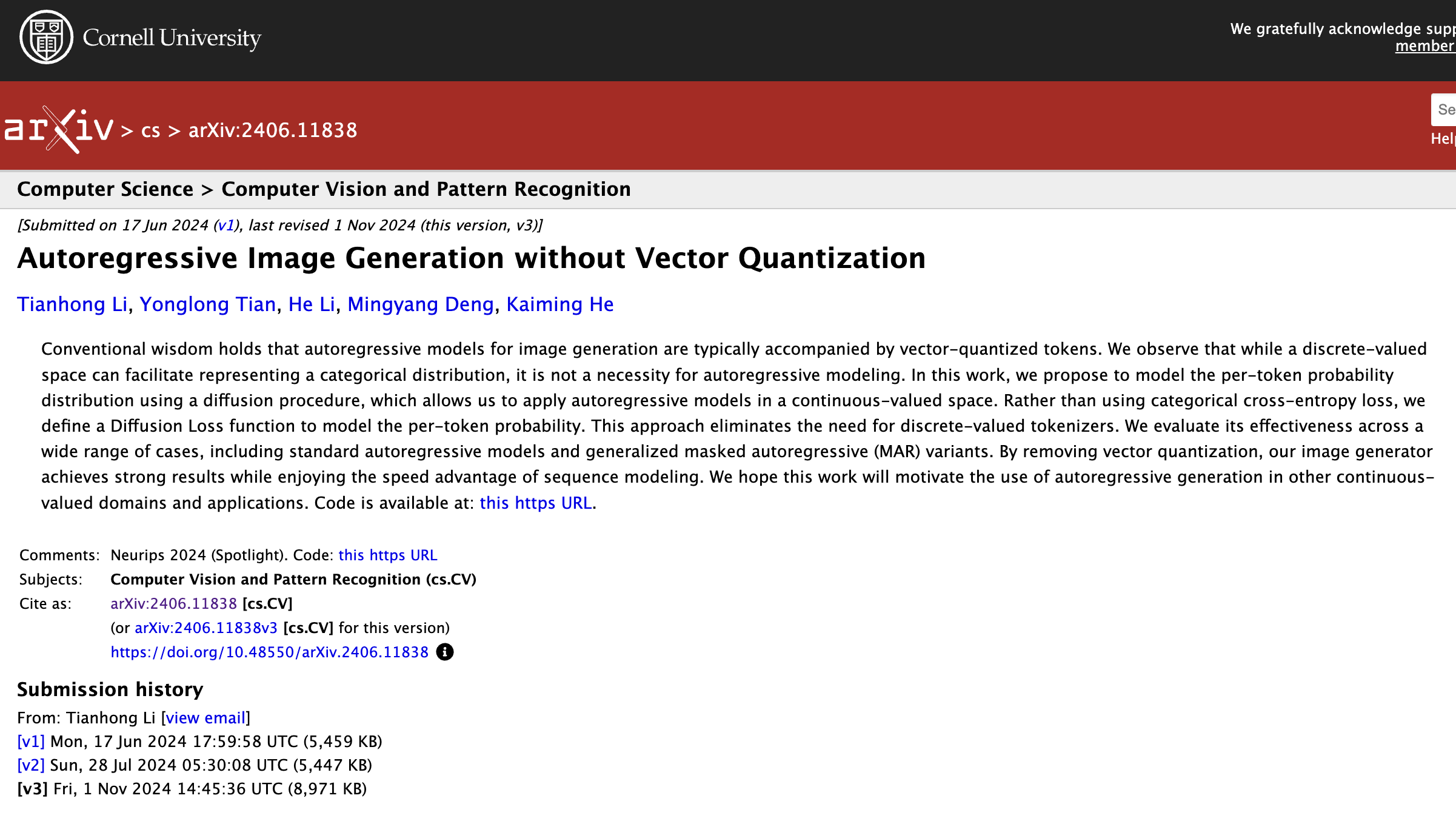} \\
\footnotesize
\textbf{User} & Please output all the text information. \\
\midrule
\footnotesize
\textbf{EVEv2.0} & Cornell University We gratefully acknowledge support member. arXiv:2406.11838 Computer Science $>$ Computer Vision and Pattern Recognition Submitted on 17 Jun 2024 (v1), last revised 1 Nov 2024 (this version, v3)) 

Autoregressive Image Generation without Vector Quantization 

Tianhong Li, Yonglong Tian, He Li, Mingyang Deng, Kaiming He

Conventional wisdom holds that autoregressive models for image generation are typically accompanied by vector-quantized tokens. We observe that while a discrete-valued space can facilitate representing a categorical distribution, it is not a necessity for autoregressive modeling. In this work, we propose to model the per-token probability distribution using a diffusion procedure, which allows us to apply autoregressive models in a continuous-valued space. Rather than using categorical cross-entropy loss, we define a Diffusion Loss function to model the per-token probability. This approach eliminates the need for discrete-valued tokenizers. We evaluate its effectiveness across a wide range of cases, including standard autoregressive models and generalized masked autoregressive (MAR) variants. By removing vector quantization, our image generator achieves strong results while enjoying the speed advantage of sequence modeling. We hope this work will motivate the use of autoregressive generation in other continuous-valued domains and applications. Code is available at: this https URL 

Comments: Neurips 2024 (Spotlight); Code: this https URL Subjects: Computer Vision and Pattern Recognition (cs.CV) Cite as: arXiv:2406.11838 [cs.CV] (or arXiv:2406.11838v3 [cs.CV] for this version) Submission history From: Tianhong Li [view email] [v1] Mon, 17 Jun 2024 17:59:58 UTC (5,459 KB) [v2] Sun, 28 Jul 2024 05:30:08 UTC (5,447 KB) [v3] Fri, 1 Nov 2024 14:45:36 UTC (8,971 KB) \\
\end{tabular}
}
\end{AIbox}
\captionof{table}{EVEv2.0 exhibits excellent OCR recognition capabilities in understanding Webpage.}
\vspace{-1mm}
\label{tab:ocr_in_web}
\end{minipage}
\end{table*}

\begin{table*}
\begin{minipage}{0.99\textwidth}
\begin{AIbox}{Example 2: Understanding OCR Information in PPT}
\centering
\scalebox{0.80}{
\begin{tabular}{l p{18cm}}
&  \includegraphics[width=16cm]{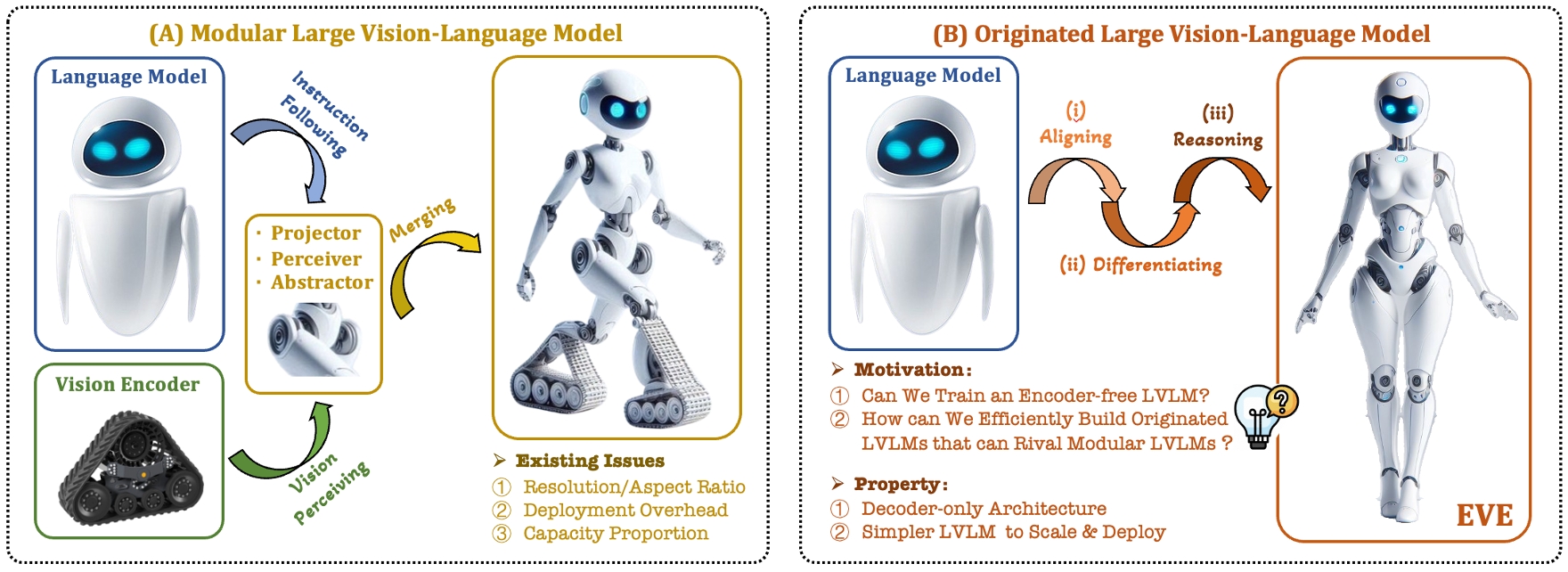} \\
\footnotesize
\textbf{User} & Please output all the text information. \\
\midrule
\footnotesize
\textbf{EVEv2.0} & (A) Modular Large Vision-Language Model (B) Originated Large Vision-Language Model Language Model Interaction Following (i) Aligning Reasoning (ii) Differentiating Vision Encoder Perceiver Merging Projector Abstractor Vision Encoder Vision Encoder Existing Issues (1) Resolution/Aspect Ratio (2) Deployment Overhead (3) Capacity Proportion Motivation: Can We Train an Encoder-free LVLM? How can We Efficiently Build Originated LVLMs that can Rival Modular LVLMs? Property: (1) Decoder-only Architecture (2) Simpler LVLM to Scale \& Deploy EVE. \\
\end{tabular}
}
\end{AIbox}
\captionof{table}{EVEv2.0 exhibits excellent OCR recognition capabilities in understanding PowerPoint.}
\vspace{-1mm}
\label{tab:ocr_in_ppt}
\end{minipage}
\end{table*}

\begin{table*}
\begin{minipage}{0.99\textwidth}
\begin{AIbox}{Example 3: Understanding Poster Content}
\centering
\scalebox{0.80}{
\begin{tabular}{l p{18cm}}
&  \includegraphics[width=16cm]{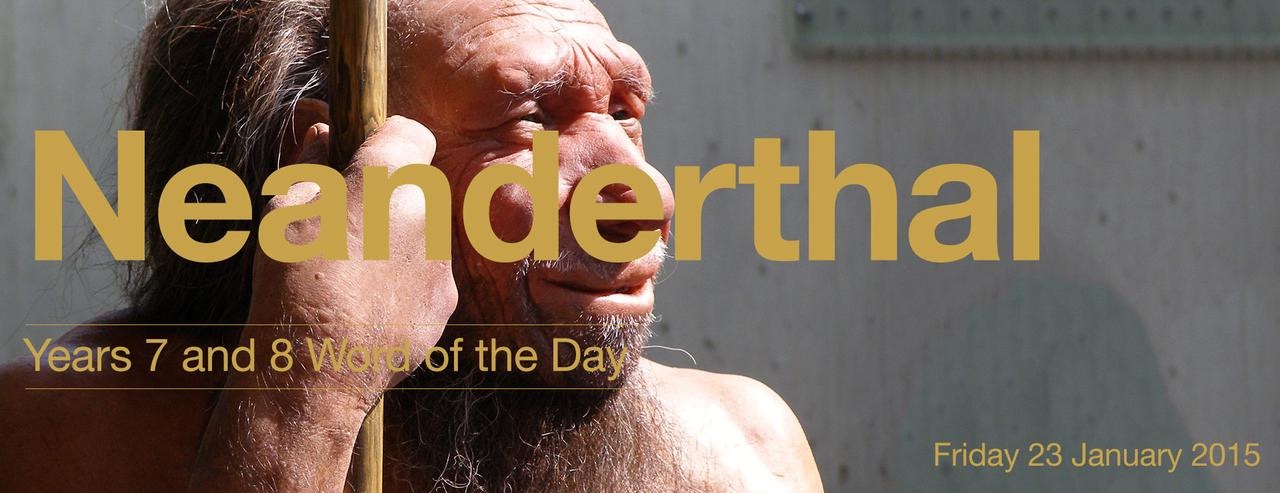} \\
\footnotesize
\textbf{User} & Please describe the image in detail. \\
\midrule
\footnotesize
\textbf{EVEv2.0} & The image is a promotional poster for "Neanderthal," which appears to be a documentary or film about Neanderthals. The title "Neanderthal" is prominently displayed in large, bold yellow letters across the center of the image. Below the title, there is additional text that reads "Years 7 and 8 Welcome of the Day," suggesting that this event might be targeted towards students in those grade levels.

The background of the poster features a close-up photograph of a man with a Neanderthal-like appearance. He has a long beard, a prominent forehead, and a strong jawline, which are characteristic features often associated with Neanderthals. The man is holding a wooden stick or pole, which adds to the prehistoric theme of the image. The background behind him is blurred, drawing focus to his face and upper body.

At the bottom right corner of the poster, there is more text indicating the date of the event: "Friday 23 January 2015." This provides a specific time frame for when the event or release of the documentary/film would have taken place.

Overall, the poster uses visual elements such as the Neanderthal figure and the title to convey information about the subject matter, while also providing essential details like the target audience and the date of the event. The design effectively captures attention and communicates key information to viewers. \\
\end{tabular}
}
\end{AIbox}
\captionof{table}{EVEv2.0 exhibits excellent visual recognition capabilities in understanding Poster.}
\vspace{-1mm}
\label{tab:poster}
\end{minipage}
\end{table*}

\begin{table*}
\begin{minipage}{0.99\textwidth}
\begin{AIbox}{Example 4: Understanding Real-world Scene}
\centering
\scalebox{0.80}{
\begin{tabular}{l p{18cm}}
&  \includegraphics[width=16cm]{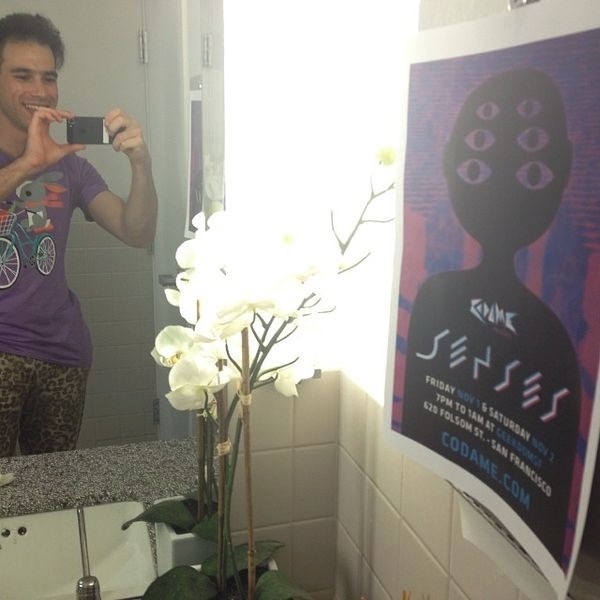} \\
\footnotesize
\textbf{User} & Please describe the image in detail. \\
\midrule
\footnotesize
\textbf{EVEv2.0} & In the image, there is a man standing in front of a bathroom mirror, taking a selfie with his cell phone. He is wearing a purple shirt and leopard print pants. The man appears to be smiling as he captures the photo.

In the background, there is a potted plant with white flowers placed on the counter near the sink. The sink is white and has a modern design. There is also a poster hanging on the wall next to the mirror. The poster features a silhouette of a person's head with several circles on it. Below the silhouette, there is text that reads "COME SETS," followed by some additional information about Friday night events at Codame.com.

The bathroom appears to be well-lit, likely due to natural light coming from a window or another source not visible in the image. The overall atmosphere seems casual and relaxed, with the man enjoying his time in the bathroom while capturing a moment through his phone camera. \\
\end{tabular}
}
\end{AIbox}
\captionof{table}{EVEv2.0 exhibits excellent recognition capabilities in the real-world scene.}
\vspace{-1mm}
\label{tab:real1}
\end{minipage}
\end{table*}

\begin{table*}
\begin{minipage}{0.99\textwidth}
\begin{AIbox}{Example 5: Understanding Real-world Scene}
\centering
\scalebox{0.80}{
\begin{tabular}{l p{18cm}}
&  \includegraphics[width=16cm]{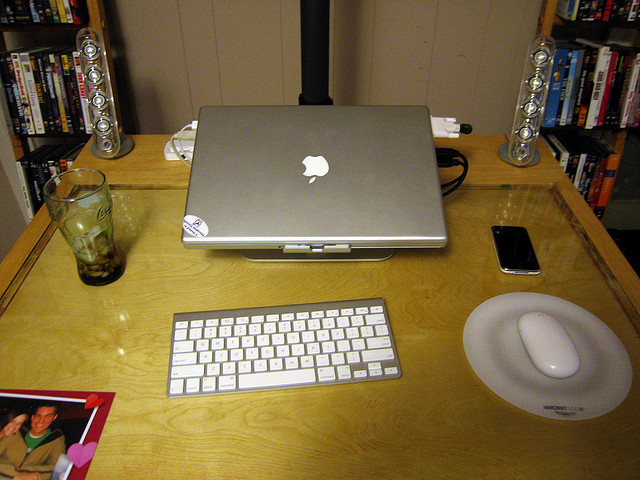} \\
\footnotesize
\textbf{User} & Please describe the image in detail. \\
\midrule
\footnotesize
\textbf{EVEv2.0} & The image depicts a well-organized workspace featuring a wooden desk with various items on it. At the center of the desk is a closed silver laptop computer, which has an Apple logo on its lid. This laptop is accompanied by a white keyboard placed directly in front of it, suggesting that the user might prefer using an external keyboard for typing.

To the right of the laptop, there is a white wireless mouse, indicating that the user also utilizes a mouse for navigation and other tasks. On the left side of the desk, there is a glass containing a beverage, possibly water or another drink, which is positioned near the edge of the desk.

In addition to these items, there are two cylindrical objects resembling speakers, one on each side of the desk. These speakers likely provide audio output for the user while working or listening to music.

Behind the desk, there is a bookshelf filled with numerous books, suggesting that the user may have a keen interest in reading or research. The presence of multiple books indicates that this workspace could be used for studying, work, or leisure activities involving reading.

Overall, the image portrays a functional and organized workspace designed for productivity and comfort, with essential computing equipment and additional accessories such as speakers and a beverage. \\
\end{tabular}
}
\end{AIbox}
\captionof{table}{EVEv2.0 exhibits excellent recognition capabilities in the real-world scene.}
\vspace{-1mm}
\label{tab:real2}
\end{minipage}
\end{table*}

\begin{table*}
\begin{minipage}{0.99\textwidth}
\begin{AIbox}{Example 6: Set-of-mark Prompting}
\centering
\scalebox{0.80}{
\begin{tabular}{l p{18cm}}
&  \includegraphics[width=16cm]{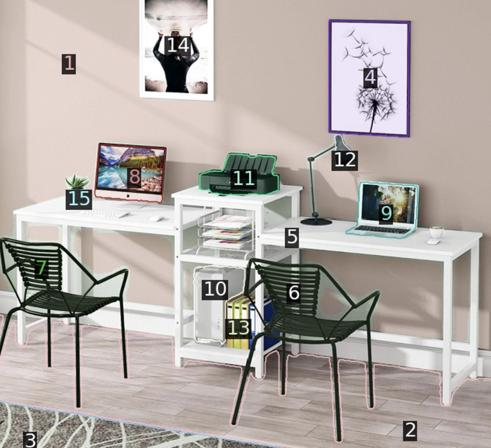} \\
\footnotesize
\textbf{User} & Here is an image with objects labeled in numbers. Please describe the objects that are marked as 9, 12 and 15. \\
\midrule
\footnotesize
\textbf{EVEv2.0} & In the image, there are three objects marked with numbers: 9, 12, and 15. 

Object 9: A laptop computer is placed on the desk. It appears to be open and ready for use.

Object 12: A lamp is situated near the desk, providing light for the workspace. It has a sleek design and complements the overall aesthetic of the room.

Object 15: A potted plant is located on the desk, adding a touch of greenery and life to the space. It contributes to creating a more pleasant and inviting atmosphere in the room. \\
\end{tabular}
}
\end{AIbox}
\captionof{table}{EVEv2.0 exhibits excellent set-of-mark prompting capabilities, ie, referring to marks when answering questions.}
\vspace{-1mm}
\label{tab:refer_to_marks}
\end{minipage}
\end{table*}

\end{appendix}

\end{document}